%% file: main.tex
\documentclass[11pt]{article}

\usepackage[preprint]{acl}

\usepackage{times}
\usepackage{latexsym}
\usepackage[T1]{fontenc}
\usepackage[utf8]{inputenc}
\usepackage{microtype}
\usepackage{inconsolata}
\usepackage{graphicx}

\usepackage{booktabs}
\usepackage{longtable}
\usepackage{array}
\usepackage{multirow}
\usepackage{amsmath}
\usepackage{amssymb}
\usepackage[table]{xcolor}
\usepackage{subcaption}
\usepackage{xspace}
\usepackage{placeins}
\usepackage{float}
\usepackage{tabularx}

\graphicspath{{figures/}}

\newcommand{\method}{\textsc{InsightAgents}\xspace}

\title{Distill, Debate, Discover: Grounding Multi-Agent Research Ideation in Graph-Distilled Insights}

\newcommand{\codeurl}{https://github.com/MichelleMa8/InsightAgents}

\author{
  \textbf{Qianran Ma}\textsuperscript{*} \quad
  \textbf{Muxin Liu}\textsuperscript{*} \quad
  \textbf{Grace Kim}\textsuperscript{*} \quad
  \textbf{Xiao Liang} \quad
  \textbf{Chris Callison-Burch} \\
  University of Pennsylvania \\[2pt]
  \texttt{\{qianranm, muxinliu, yeeunk\}@engineering.upenn.edu}
}

\begin{document}
\maketitle
\begingroup
\renewcommand\thefootnote{*}
\footnotetext{Equal contribution.}
\renewcommand\thefootnote{}
\footnotetext{Code available at \url{\codeurl}.}
\endgroup

\begin{abstract}
\input{sections/abstract}
\end{abstract}

\input{sections/introduction}
\input{sections/related_work}

\input{sections/method}
\input{sections/experiments}
\input{sections/results}

\input{sections/analysis}
\input{sections/conclusion}

\FloatBarrier
\clearpage
\clearpage
\input{sections/limitations}

\input{sections/ethics}

\input{sections/acknowledgments}

\bibliography{references}

\clearpage
\appendix
\input{appendix/appendix}

\end{document}

%% file: sections/abstract.tex

Large language models (LLMs) are increasingly used to support research idea generation and scientific discovery.
Multi-agent debate offers a collaborative approach to this task by bringing together diverse perspectives to develop research ideas.
However, a key challenge is how to equip agents with distinct and meaningful expertise to guide their contributions.
Inspired by how researchers draw on accumulated research experience during collaboration, we introduce \method, a framework that constructs literature-grounded agent expertise by distilling insights from graphs of related papers.
Guided by insights that trace evolving research approaches, reveal underlying assumptions, and identify unresolved challenges, within a line of research, agents jointly identify a research problem and propose, challenge, and refine potential approaches through debate to develop a research proposal.
We evaluate \method against three research idea generation frameworks on 100 NLP research topics through pairwise comparisons by two LLM judges.
In both online and offline knowledge settings, \method achieves the highest average score across the five evaluation dimensions under both judges.

%% file: sections/introduction.tex
\section{Introduction}
\label{sec:intro}

Recent research frameworks use large language models (LLMs) to automate multiple stages of scientific work, from literature review and idea development to experimentation and paper writing \citep{yamada-etal-2025-ai-scientist-v2,schmidgall-etal-2025-agent,tang-etal-2025-ai-researcher}.
The quality of an initial research idea can shape subsequent research outcomes \citep{si-etal-2025-ideation-execution}.
We therefore focus on generating research ideas grounded in relevant literature through multi-agent collaboration.

Prior work has explored different ways of using literature and agent collaboration for research idea generation.
ResearchAgent retrieves related papers and entity knowledge to support idea generation, with reviewing agents providing feedback \citep{baek-etal-2025-researchagent}.
Chain of Ideas organizes papers into development chains and uses their trends to guide idea generation \citep{li-etal-2025-chain-ideas}.
Multi-agent approaches also vary in how they initialize agents: VirSci uses real researcher profiles, ResearchTown constructs researcher profiles from papers, and Ueda et al. explore assigning domain-specific personas to agents \citep{su-etal-2025-many,yu-etal-2025-researchtown,ueda-etal-2025-exploring}.
These works illustrate literature-guided idea generation and profile- or persona-based agent initialization.
However, few approaches initialize debate agents with insights distilled from different lines of research.

Researchers develop expertise by tracing how lines of research evolve across related papers, learning from prior approaches, their assumptions, and unresolved challenges \citep{boote2005scholars, li-etal-2025-chain-ideas}.
Inspired by this process, we equip agents with research perspectives grounded in the literature.
We define a \emph{research insight} as a synthesis of how a line of research has evolved, capturing its key approaches, underlying assumptions, supporting findings, and unresolved tensions.

\begin{figure*}[t]
    \centering
    \includegraphics[width=\textwidth]{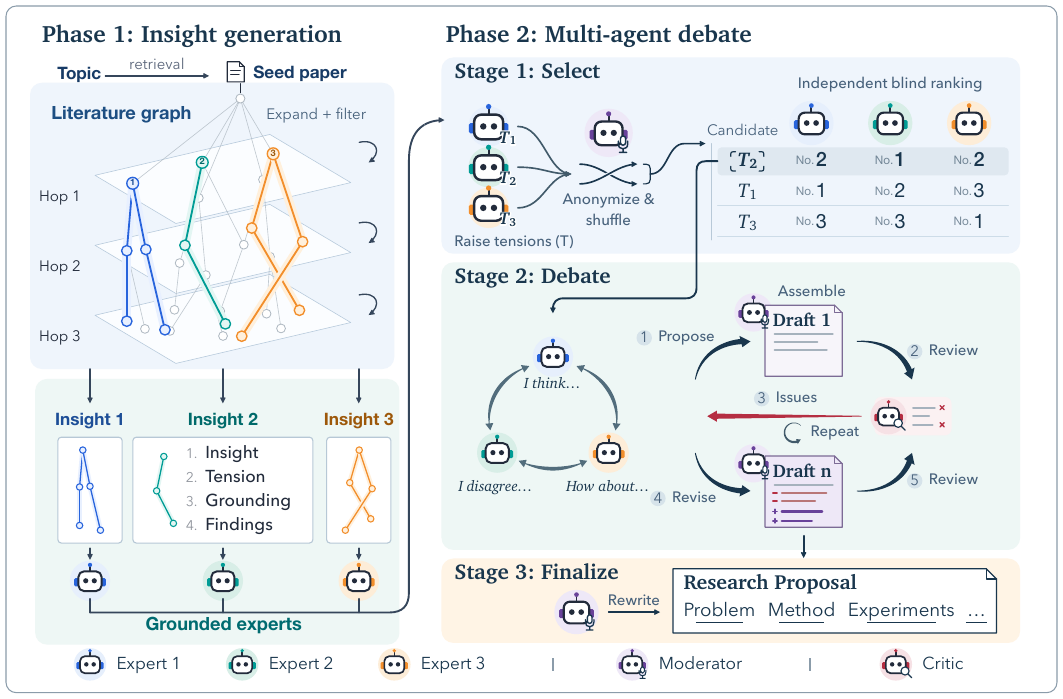}
    \caption{Overview of \method. Insights distilled from graph-structured literature define agent expertise, with each insight linked to its supporting evidence. The agents use these perspectives to develop a research proposal through multi-agent debate.}
    \label{fig:framework}
\end{figure*}

To support research idea generation through literature-grounded multi-agent debate, we introduce \method, a framework that uses distilled research insights to initialize expert agents for multi-agent debate.
Given a research topic, \method builds a graph of related papers and distills multiple research insights, each linked to supporting findings and source papers.
Each expert agent is initialized with one insight and its supporting findings, which guide the agents as they identify a research problem and propose, challenge, and refine approaches through debate.
A moderator coordinates the discussion, while a critic provides feedback on draft proposals to guide further refinement.
Together, the agents develop a research proposal specifying the problem and motivation, proposed method, and evaluation plan.
The overall process is illustrated in Figure~\ref{fig:framework}.

We evaluate \method on 100 NLP research topics, comparing it with ResearchAgent, AI Scientist-v2, and Chain of Ideas.
Two LLM judges assess the proposals through pairwise comparisons on novelty, significance, feasibility, clarity, and expected effectiveness.
\method achieves the highest average score among the compared methods under both judges.

Our contributions are as follows:
\begin{itemize}
    \setlength{\itemsep}{2pt plus 1pt minus 0.5pt}
    \setlength{\parsep}{2pt plus 1pt minus 0.5pt}
    \setlength{\parskip}{\parsep}
    \item We define \emph{research insights} as syntheses of how lines of research evolve and distill them from a topic-specific paper graph, together with supporting findings, to initialize expert agents.
    \item We introduce \method, a framework that brings these experts together to identify research problems and propose and refine approaches through multi-agent debate, ultimately producing a research proposal.
    \item We evaluate \method on 100 NLP research topics, where it achieves the highest average score among the compared methods in two knowledge settings under two LLM judges.
\end{itemize}

%% file: sections/related_work.tex
\section{Related Work}
\label{sec:related}

\input{tables/feature_comparison}

\paragraph{Literature-grounded research idea generation}
Prior work has explored different ways of grounding research idea generation in scientific literature.
SciMON, ResearchAgent, and AI Scientist-v2 retrieve relevant papers to inform idea generation and refinement \citep{wang-etal-2024-scimon,baek-etal-2025-researchagent,yamada-etal-2025-ai-scientist-v2}.
Other methods organize literature into chains or graphs to capture research developments and relationships among scientific knowledge \citep{li-etal-2025-chain-ideas,gao-etal-2026-graphs,li-etal-2026-graph2idea,xu-etal-2026-idea2story}.
While retrieval-based methods typically do not explicitly trace how research evolves, structure-aware methods primarily use these structures to guide idea generation, rather than explore how different lines of research might challenge or complement one another.
In contrast, \method grounds multi-agent debate in insights distilled from distinct lines of research, allowing these perspectives to interact during idea development.

\paragraph{Multi-agent research ideation}
Multi-agent debate allows LLM agents to exchange and critique ideas, encouraging diverse perspectives and iterative refinement \citep{du-etal-2024-improving,liang-etal-2024-encouraging}.
Existing scientific collaboration frameworks introduce agent diversity through functional roles or domain-specific personas \citep{ghafarollahi-buehler-2025-sciagents,ueda-etal-2025-exploring}, researcher profiles \citep{su-etal-2025-many,yu-etal-2025-researchtown}, or different knowledge backgrounds \citep{yu-etal-2025-dynamic}.
However, these forms of diversity do not equip agents with distinct and topic-specific expertise grounded in prior research findings and assumptions.
In contrast, \method initializes each expert with insights distilled from a different line of research, providing a concrete and literature-grounded basis for proposing and challenging ideas during debate.
Table~\ref{tab:feature-comparison} summarizes these differences across research idea generation frameworks.

%% file: tables/feature_comparison.tex
\begin{table*}[t]
\centering
\small
\setlength{\tabcolsep}{3pt}
\renewcommand{\arraystretch}{1.08}
\newcommand{\fcYes}{\ensuremath{\checkmark}}
\newcommand{\fcNo}{\textemdash}
\newcommand{\fcAvYes}{\textcolor{black!65}{\fcYes}}
\newcommand{\fcAvNo}{\textcolor{black!65}{\fcNo}}
\newcommand{\fcAvPartial}{\textcolor{black!65}{\ensuremath{\sim}}}
\begin{tabular*}{\textwidth}{@{\extracolsep{\fill}}lcc@{\hspace{1.3em}}ccccl@{}}
\toprule
& \multicolumn{2}{c}{\textcolor{black!65}{\textbf{Availability \& input}}}
& \multicolumn{5}{c}{\textbf{Method features}} \\
\cmidrule(lr){2-3}\cmidrule(l){4-8}
Framework
& \textcolor{black!65}{\shortstack{Open\\source}}
& \textcolor{black!65}{\shortstack{Topic\\control}}
& \shortstack{Lit.\\retrieval}
& \shortstack{Lit.\\graph}
& \shortstack{Insight\\extraction}
& \shortstack{Multi-agent\\debate}
& \shortstack[l]{Agent expertise\\source} \\
\midrule
SciMON \citeyearpar{wang-etal-2024-scimon}
& \fcAvYes & \fcAvPartial & \fcYes & \fcYes & \fcNo & \fcNo & \fcNo \\
\cellcolor{black!7}ResearchAgent \citeyearpar{baek-etal-2025-researchagent}
& \fcAvYes & \fcAvPartial & \fcYes & \fcYes & \fcNo & \fcNo & \fcNo \\
\cellcolor{black!7}AI Scientist-v2 \citeyearpar{yamada-etal-2025-ai-scientist-v2}
& \fcAvYes & \fcAvYes & \fcYes & \fcNo & \fcNo & \fcNo & \fcNo \\
\cellcolor{black!7}Chain of Ideas \citeyearpar{li-etal-2025-chain-ideas}
& \fcAvYes & \fcAvYes & \fcYes & \fcYes & \fcYes & \fcNo & \fcNo \\
SciAgents \citeyearpar{ghafarollahi-buehler-2025-sciagents}
& \fcAvYes & \fcAvPartial & \fcYes & \fcYes & \fcNo & \fcYes & Functional roles \\
VirSci \citeyearpar{su-etal-2025-many}
& \fcAvYes & \fcAvNo & \fcYes & \fcNo & \fcNo & \fcYes & Author profiles \\
IDVSCI \citeyearpar{yu-etal-2025-dynamic}
& \fcAvNo & \fcAvNo & \fcYes & \fcNo & \fcNo & \fcYes & Scientist backgrounds \\
Graphs of Research \citeyearpar{gao-etal-2026-graphs}
& \fcAvNo & \fcAvPartial & \fcYes & \fcYes & \fcNo & \fcNo & \fcNo \\
Graph2Idea \citeyearpar{li-etal-2026-graph2idea}
& \fcAvNo & \fcAvPartial & \fcYes & \fcYes & \fcNo & \fcNo & \fcNo \\
\midrule
\method
& \textcolor{black!65}{\fcYes} & \fcAvYes & \fcYes & \fcYes & \fcYes & \fcYes & Topic-specific insights \\
\bottomrule
\end{tabular*}
\caption{Comparison with research idea generation frameworks.
Shaded framework names indicate the three baselines used in our experiments.
\fcYes: documented support; \fcNo: not documented or no official code located.
Open source denotes public official code.
For topic control, \fcYes denotes direct topic-to-idea input; $\sim$ denotes indirect or constrained control.
Literature graphs include chains and citation or knowledge graphs.
Insight extraction denotes intermediate trends, patterns, gaps, or transferable mechanisms.
Debate requires agents to exchange and respond to arguments, beyond independent reviews.
Agent expertise source describes what defines the expertise of debating agents; \fcNo{} marks frameworks without debate.}
\label{tab:feature-comparison}
\end{table*}

%% file: sections/method.tex
\section{Method}
\label{sec:method}

\method transforms insights distilled from graph-structured literature into agent expertise for research idea generation.
As shown in Figure~\ref{fig:framework}, the framework comprises two components: graph-based expertise construction and multi-agent debate.

\subsection{Constructing Agent Expertise}
\label{sec:expertise}

\paragraph{Paper graph construction}
We construct a paper graph $G=(V,E)$ to organize related literature for insight distillation.
The seed node $v_0$ is either a user-specified paper or one paper retrieved for the input topic.
Starting from $v_0$, we expand each node $v$ through Semantic Scholar's recommendation API, which returns a one-hop candidate set $\mathcal{C}(v)$ of papers similar to $v$.
Among new candidates $\mathcal{C}(v)\setminus V$, an LLM selects papers relevant to both the topic and $v$, favoring distinct methods, settings, or failure modes.
Candidates already in $V$ are linked directly to $v$, while selected new papers are added to $V$ and connected to $v$ through new edges in $E$.
We orient recommendation links from older to newer papers by publication date.
We recursively expand newly added nodes for three rounds, yielding up to three expansion levels beyond $v_0$.
The resulting graph collects related papers to the topic for methodological comparison and insight distillation.

\paragraph{Insight distillation}
Given the paper graph $G=(V,E)$, we prompt an LLM to jointly distill three research insights $\mathcal{I}=\{I_1,I_2,I_3\}$ using the paper titles and abstracts stored in its nodes and the relationships between papers represented by its edges.
The LLM first groups papers by the type of research they pursue, such as measurement, training, or dataset construction, and selects three distinct lines of work, each supported by at least two papers.
For each line, it synthesizes recurring challenges, the evolution of research approaches, underlying assumptions, and unresolved tensions into an insight $I_k$.
Each insight is grounded in a supporting subgraph $G_k=(V_k,E_k)$ and a set of findings $\mathcal{F}_k$, retaining specific methods, benchmarks, and reported results linked to their source papers.

\paragraph{Expertise formation}
For each research insight $I_k$, we initialize an expert agent $A_k$ with a system prompt containing the insight, its supporting findings $\mathcal{F}_k$, and the titles and abstracts of papers in $V_k$.
The insight provides the agent with a perspective on a line of research, while the findings and source abstracts ground this perspective in specific methods, benchmarks, and reported results.
During debate, the agent is instructed to reason from this perspective while drawing on its broader knowledge to propose and evaluate research approaches.

\subsection{Research Idea Generation through Multi-Agent Debate}
\label{sec:debate}

The debate involves three roles: expert agents, a moderator, and a critic.
Each expert agent $A_k$ draws on its insight $I_k$ and supporting findings to identify research problems, propose approaches, and evaluate the shared proposal.
The moderator coordinates the discussion and integrates expert contributions into a common draft.
To encourage critical examination rather than premature consensus \citep{pitre-etal-2025-consensagent}, the critic acts as an independent reviewer, identifying weaknesses in the current draft.
It receives the draft and the latest discussion summary, without access to the full debate history, and provides feedback to guide subsequent expert revisions.
Together, these roles develop one research proposal through the following three stages.

\paragraph{Stage 1: Problem selection}
Each expert proposes a research problem based on its insight.
The moderator consolidates these proposals into three anonymous candidates, which are shuffled before the experts rank them independently.
Rankings are aggregated using Borda scoring \citep{young1974axiomatization}, with 3, 2, and 1 points for first, second, and third place, respectively.
The moderator summarizes the highest-scoring problem for subsequent debate, choosing among the tied candidates only when necessary.

\paragraph{Stage 2: Proposal development and debate}
For the selected problem, experts develop a shared proposal through three types of rounds: proposal, refinement, and convergence.
In the proposal round, each expert proposes an approach informed by its expertise and, where useful, broader knowledge.
The moderator identifies the strongest direction supported across the proposals, combining complementary approaches, and produces an initial draft.
The critic then reviews the draft for scientific contribution, soundness, consistency, feasibility, and clarity.

In subsequent refinement rounds, experts respond to the critic's feedback, proposing improvements to the method, claims, or evaluation.
The moderator updates the draft and summarizes resolved, unresolved, and newly raised issues.
The critic reviews each revised draft and identifies issues to be addressed in the next round.

In the final convergence round, experts address remaining concerns and state their final positions without introducing new research directions.

\paragraph{Stage 3: Final synthesis}
After the debate, the moderator integrates the latest draft and the experts' final positions into a final research proposal.
It retains the selected research problem, incorporates the methods the experts agreed on during debate, and reports unresolved concerns as limitations.
The final proposal specifies the problem and motivation, method, experiment, anticipated novelty, and potential impact.

%% file: sections/experiments.tex
\section{Experimental Setup}
\label{sec:experiments}

\subsection{Pipeline Setup}
\label{sec:implementation}

\input{tables/knowledge_settings}

\paragraph{Knowledge settings}
We evaluate \method under two knowledge settings that reflect different ways of accessing prior research: retrieving papers on demand from an evolving body of literature and exploring a fixed and structured knowledge resource.

In the \textbf{online} setting, we dynamically construct a paper graph using the Semantic Scholar API, as described in Section~\ref{sec:method}.
In the \textbf{offline} setting, we use a preconstructed scientific contribution graph \citep{jansen-2026-scientific-contribution-graph} and map its contribution nodes to their source papers, preserving the prerequisite relationships between contributions as paper-level connections.
Table~\ref{tab:knowledge-settings} summarizes the key differences between the two settings.
Both use the same insight distillation, expertise formation, and debate procedures, while graph retrieval and expansion differ.
Evaluating both settings allows us to examine whether \method remains effective across different knowledge sources and graph structures.
Appendix~\ref{sec:appendix-offline} details the offline graph retrieval process and the evidence provided to agents.

Since the settings differ in their underlying corpora, graph construction procedures, edge semantics, and seed selection, we treat them as two configurations of \method rather than a controlled comparison of knowledge sources.

\paragraph{Topic selection}
We evaluate our framework on the same set of 100 NLP research topics in both knowledge settings.
To construct the candidate topic pool, we randomly sample papers from the 2025 main proceedings and available Findings volumes of ACL, EMNLP, NAACL, and COLING.
We use GPT-5.6 Luna to derive research topics from their titles and abstracts, exclude unsuitable source papers (e.g., surveys), and remove semantic duplicates.
From the resulting pool, we randomly select 100 evaluation topics spanning 20 research areas.
Appendix~\ref{sec:appendix-topic-selection} describes the sampling, abstraction, filtering, and deduplication rules and reports the area distribution.

\paragraph{Generation configuration}
We use GPT-5.6 Luna for insight extraction and all debate roles in \method.
Each topic is processed by three insight-based experts, a moderator, and a critic.
Problem selection uses two rounds, followed by three debate rounds and a final rewrite that produces one research proposal.
Graph construction parameters and generation budgets are provided in Appendix~\ref{sec:appendix-implementation}.
Prompt templates are provided in Appendix~\ref{sec:appendix-prompts}.

\subsection{Baselines}
\label{sec:baselines}

We select baselines from the frameworks in Table~\ref{tab:feature-comparison} that have public code and can generate ideas for a given topic, either from the topic itself or from a single seed paper.
Three frameworks meet both criteria: ResearchAgent \citep{baek-etal-2025-researchagent}, AI Scientist-v2 \citep{yamada-etal-2025-ai-scientist-v2}, and Chain of Ideas \citep{li-etal-2025-chain-ideas}.
AI Scientist-v2 and Chain of Ideas take the topic directly; ResearchAgent takes the seed paper selected by \method in the online setting, with the topic text added to its problem prompts.
The multi-agent frameworks in Table~\ref{tab:feature-comparison} do not meet these criteria.
VirSci forms a research team that chooses its own topic, SciAgents generates hypotheses from keywords in a fixed concept graph, and IDVSCI has no public code.
Adapting them to a given topic would change their task, so we do not include them as baselines.

We evaluate all three baselines using the same 100 research topics and generation model (GPT-5.6 Luna) as \method.
We use the official implementations with default pipeline settings, except for the adaptations detailed in Appendix~\ref{sec:appendix-baselines}.
Since we compare complete pipelines, we do not control for generation costs across methods.

\subsection{Evaluation Protocol}
\label{sec:evaluation}

\paragraph{LLM evaluation}
We adopt pairwise evaluation, as prior studies suggest that LLM judges align more closely with human judgments when comparing outputs than when assigning absolute scores \citep{liusie-etal-2024-llm,liu-etal-2024-aligning}.
Following Chain of Ideas \citep{li-etal-2025-chain-ideas}, we evaluate research ideas along five dimensions: novelty, significance, feasibility, clarity, and expected effectiveness.
Table~\ref{tab:prompts-evaluation} provides detailed definitions of these dimensions and the evaluation prompt.

We use GPT-5.6 Sol and Claude Opus 4.8 as judges, with the same evaluation prompt across methods, and report their results separately.
For each topic, the judge compares every pair of methods and selects a winner or a tie on each dimension, with a brief rationale.
To reduce position and stylistic biases, we evaluate each pair in both presentation orders and instruct the judge to focus on content rather than surface features such as length and formatting \citep{feuer-etal-2025-style}.

\paragraph{Human--LLM agreement}
To assess agreement between human and LLM judgments, two NLP researchers independently evaluate proposals on 10 research topics selected based on their familiarity with the topics.
For each topic, annotators rank the four proposals from \method (online setting) and the three baselines along the same five evaluation dimensions, allowing ties.
Annotators receive the same criteria and instructions as the LLM judges, and method identities are hidden during annotation.
We derive pairwise preferences from these rankings and measure both inter-annotator and human--LLM agreement.

%% file: tables/knowledge_settings.tex
\begin{table}[t]
\centering
\footnotesize
\setlength{\tabcolsep}{3pt}
\renewcommand{\arraystretch}{1.15}
\begin{tabularx}{\columnwidth}{
    @{}
    l
    >{\raggedright\arraybackslash}X
    >{\raggedright\arraybackslash}X
    @{}
}
\toprule
\textbf{Aspect} & \textbf{Online} & \textbf{Offline} \\
\midrule
Source
& Semantic Scholar API
& Contribution graph \\

Construction
& On demand
& From prebuilt graph \\

Nodes
& Papers
& Papers \\

Edges
& Recommendations
& Prerequisite relations \\

Direction
& Older $\rightarrow$ newer
& Prerequisite $\rightarrow$ dependent \\

\bottomrule
\end{tabularx}
\caption{Comparison of the two knowledge settings used by \method. In the offline setting, contribution nodes are mapped to their source papers.}

\label{tab:knowledge-settings}
\end{table}

%% file: sections/results.tex
\section{Results}
\label{sec:results}

\begin{figure*}[!t]
    \centering
    \includegraphics[width=\textwidth]{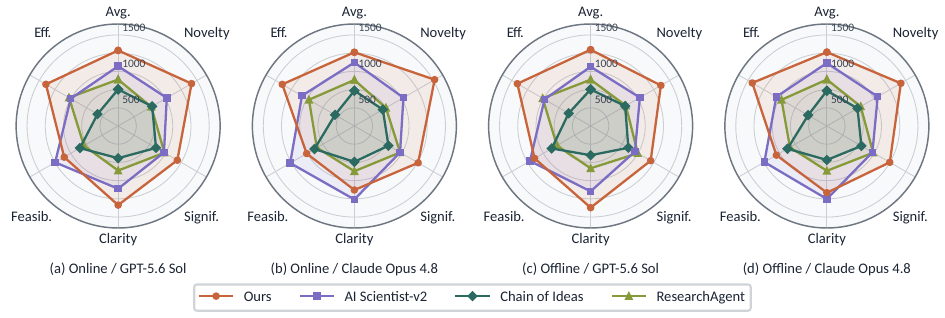}
    \caption{Elo profiles on 100 topics per setting: (a--b) online and (c--d) offline, each evaluated by GPT-5.6 Sol and Claude Opus 4.8. Ours denotes \method; Avg., Signif., Feasib., and Eff. denote average Elo, significance, feasibility, and expected effectiveness. All panels use the same radial scale (250--1,650; higher is better).}
    \label{fig:results-radar}
\end{figure*}

\paragraph{Main results}
We aggregate pairwise judgments into Elo ratings for each evaluation dimension, knowledge setting, and judge, with equal weights across the five dimensions for the average Elo.
Figure~\ref{fig:results-radar} presents the comparative evaluation results.

\method achieves the highest average Elo in all four combinations of knowledge setting and judge, followed by AI Scientist-v2.
Both judges rank \method first in novelty, significance, and expected effectiveness under both settings.

Feasibility represents the main trade-off, with AI Scientist-v2 ranking first and \method second across both settings and judges.
Appendix~\ref{sec:appendix-feasibility} further examines the judges' rationales for this gap in the online setting.
Clarity rankings vary by judge: GPT-5.6 Sol ranks \method first, while Claude Opus 4.8 favors AI Scientist-v2, placing \method second in both settings.
Detailed Elo ratings and direct pairwise comparison scores are provided in Appendix~\ref{sec:appendix-results}.

The consistent results across both settings suggest that \method can effectively leverage different forms of literature-grounded knowledge without relying on a specific graph construction procedure.

\begin{figure}[!t]
    \centering
    \includegraphics[width=\columnwidth]{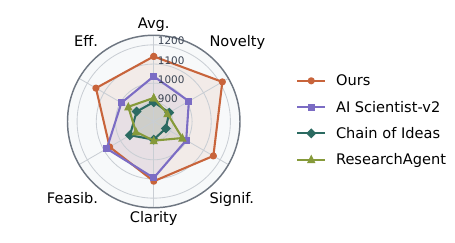}
    \caption{Human Elo ratings on 10 online topics, derived from rankings across all four methods. Ours denotes \method. Axes follow Figure~\ref{fig:results-radar}; the radial scale is 800--1,250.}
    \label{fig:human-radar}
\end{figure}

\paragraph{Human--LLM agreement}
Figure~\ref{fig:human-radar} shows that \method also achieves the highest average Elo on the human-evaluated topics, with the largest margins over AI Scientist-v2 in novelty, significance, and expected effectiveness.
The two methods receive similar ratings for clarity and feasibility, with \method slightly ahead on clarity and AI Scientist-v2 on feasibility.
These results support the main pattern of the LLM evaluation on this subset.

\begin{figure}[!t]
    \centering
    \includegraphics[width=\columnwidth]{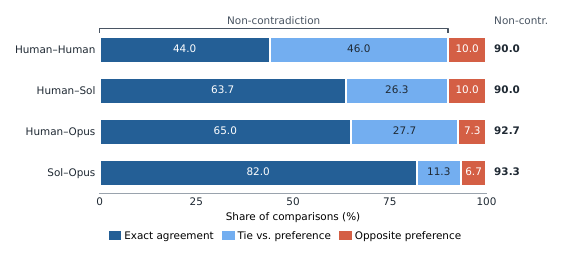}
    \caption{Agreement on \method--baseline pairs over 150 topic--dimension--pair instances from 10 online topics. Each bar splits one comparison into exact agreement (same winner or both ties), tie vs.\ preference (one judge returns a tie and the other does not), and opposite preference. Non-contradiction is the sum of the two blue segments. Sol and Opus denote GPT-5.6 Sol and Claude Opus 4.8.}
    \label{fig:human-agreement}
\end{figure}

We compare human and LLM judgments on \method--baseline pairs using two measures.
\emph{Exact agreement} requires both judges to select the same winner or both declare a tie.
\emph{Non-contradiction} counts all cases except those where the two judges prefer opposite methods, including cases where both select the same winner, both declare a tie, or one declares a tie while the other selects a winner.
Figure~\ref{fig:human-agreement} shows that opposing preferences account for only 7.3--10.0\% of human--LLM comparisons, yielding non-contradiction rates of 90.0--92.7\%, close to the 93.3\% between the two LLM judges.
Exact agreement is lower (63.7--65.0\%), with most disagreements involving ties rather than opposing preferences.
Human annotators show a similar pattern, with tie--preference disagreements (46.0\%) exceeding exact agreement (44.0\%).
Appendix~\ref{sec:appendix-human-evaluation} provides confidence intervals and dimension-level results.

%% file: sections/analysis.tex
\section{Analysis}
\label{sec:analysis}

\subsection{Ablation Study}
\input{tables/ablation}

\subsection{Generation Process Analysis}
\label{sec:generation-process-analysis}
\input{sections/generation_process_analysis}

\subsection{Comparison with Other Frameworks}
\label{sec:framework-comparison}
\begin{figure}[!t]
    \centering
    \includegraphics[width=\columnwidth]{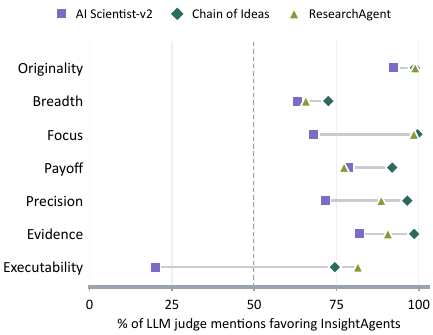}
    \caption{LLM judges' preferences by proposal property, based on their stated reasoning. Each point shows the percentage of explanations favoring \method over a baseline among those discussing that property. Results pool both LLM judges, GPT-5.6 Sol and Claude Opus 4.8 (3,000 explanations each), in the online setting. Values above 50\% (dashed line) favor \method. See Appendix~\ref{sec:appendix-rationales} for details.}
    \label{fig:rationale-column}
\end{figure}

To understand why LLM judges prefer \method, we analyze the reasoning they provide with their pairwise judgments. We use Claude Opus 5.5 to code 6,000 such explanations from the online setting, 3,000 from each of the two LLM judges, across seven proposal properties: originality (new framing or mechanisms), breadth (generalizability), focus (one clear research question), payoff (perceived value of the result), precision (specified methods and success criteria), evidence (controls that rule out alternative explanations), and executability (feasibility given available resources). The detailed descriptions of each property are included in Appendix~\ref{sec:appendix-rationales}.

Figure~\ref{fig:rationale-column} shows that judges' reasoning discussing originality or evidence favors \method across all three baselines (81.9–98.9\%). Judges highlight how its proposals reframe research problems and use experimental controls to test their claims. Against Chain of Ideas and ResearchAgent, they also favor its focus and precision. The main exception is executability against AI Scientist-v2, where 80.1\% of relevant rationales favor the baseline. Judges cite the scale of \method's proposed studies, human annotation, and custom experimental setups as obstacles.  

%% file: tables/ablation.tex
We ablate three steps of \method: insight distillation, retrieval grounding, and multi-agent debate. Each variant changes one step and keeps the rest of the pipeline fixed.

\emph{w/o insight} replaces each expert’s insight $I_k$ and findings $\mathcal{F}_k$ with the titles and abstracts of its supporting papers $V_k$. The paper graph $G$ is unchanged.
\emph{w/o retrieval} removes retrieval that the generation model writes three insights and their findings from its own knowledge, in the same format and length as \method.
\emph{w/o debate} provides all three insights to a single agent, which selects a research problem, drafts a proposal, and self-reviews and revises it using the same five criteria as the critic in the full system.
This is completed in a single response without multi-round discussion or a separate critic, followed by the same final rewrite step to maintain consistent output format and length.

We run each variant on 30 topics from the online setting with the same generation model. Claude Opus 4.8 compares each variant’s proposal against \method’s on every dimension. We report each variant's win rate against \method, counting ties as half a win.
Scores below 50\% indicate that the removed component benefits \method.

\begin{table}[H]
    \centering
    \footnotesize
    \setlength{\tabcolsep}{3.5pt}
    \caption{Ablation results against \method, which scores 50 by construction. Scores below 50 indicate that the removed step benefits the full framework.}
    \label{tab:ablation}
    \begin{tabular}{lcccccc}
        \toprule
        \textbf{Variant} & \textbf{Nov.} & \textbf{Sig.} & \textbf{Feas.} & \textbf{Clar.} & \textbf{Eff.} & \textbf{Avg.} \\
        \midrule
        \method              & 50.0 & 50.0 & 50.0 & 50.0 & 50.0 & 50.0 \\
        \quad w/o insight    & \textbf{41.7} & \textbf{43.3} & 55.0 & 51.7 & \textbf{45.0} & \textbf{47.3} \\
        \quad w/o retrieval  & \textbf{43.3} & \textbf{48.3} & 51.7 & 53.3 & 50.8 & \textbf{49.5} \\
        \quad w/o debate     & \textbf{19.2} & \textbf{40.0} & 51.7 & \textbf{42.5} & \textbf{11.7} & \textbf{33.0} \\
        \bottomrule
    \end{tabular}
\end{table}

Table~\ref{tab:ablation} shows that insight distillation and retrieval both improve novelty and significance without reducing effectiveness, because they give the experts more context about prior work.
w/o insight and w/o retrieval score slightly higher on feasibility and clarity because they propose simpler ideas.
Counting datasets, baselines, experimental designs, and human annotations, they propose 8.6 and 7.5 experimental components, against 9.1 for \method.
w/o retrieval also cites fewer and older works: 1.9 works from 2021 on average, against 3.6 from 2026 for \method.
The judge knows this older work well and finds these proposals easier to follow.
As a result, w/o retrieval reaches an average score close to \method’s, but its ideas are poorly grounded and partly favored for their familiarity to the judge.
\method instead prioritizes pushing novelty and significance over ease of execution (Appendix~\ref{sec:appendix-feasibility}).
Removing debate leads to the largest overall performance drop (33.0), particularly in novelty (19.2) and expected effectiveness (11.7), while feasibility remains nearly unchanged (51.7).
This suggests that multi-agent discussion and iterative refinement help develop more novel and potentially effective research proposals.

%% file: sections/generation_process_analysis.tex

\paragraph{Evidence diversity and use}
We first examine how literature-grounded expertise is distributed and used across 100 online topics.
Among 300 within-topic expert pairs, 289 share no supporting papers, indicating largely non-overlapping evidence sets.
In the first proposal round, 299 of 300 expert proposals reference at least one paper from their own supporting sets, with an average of 4.3 papers per proposal.
In 67 of 100 topics, the final proposal's \textit{builds on} field references evidence from at least two experts.
These statistics characterize how distinct sources of evidence enter the proposal development process.

\paragraph{Case study: Diversity of research insights}
We examine the topic \textit{Personalized dialogue generation through evolving user-specific memory} to illustrate the distinct research perspectives distilled from the same paper graph.
Expert~A (\textit{Memory Utilization Evaluation}) focuses on the gap between retrieving user memories and appropriately integrating them into responses \citep{sumida-etal-2026-memuse}.
Expert~B (\textit{Evolving Memory Architecture}) emphasizes how memory systems represent evolving evidence, including temporal changes, contradictions, and contextual validity \citep{choi-etal-2026-pgmem,liu-etal-2026-emir2,yacoubi-etal-2026-memorylace}.
Expert~C (\textit{Continual Personalization Alignment}) focuses on the challenges of adapting to evolving user preferences while avoiding irrelevant personalization, preference narrowing, and sycophancy \citep{wang-etal-2026-hidden-costs}.
These insights provide complementary perspectives on evaluating memory use, representing evolving memory states, and regulating personalization.

\paragraph{Case study: Expertise-guided debate and refinement}
To illustrate how experts interact during debate, we examine one of the issues they discussed: how to control the influence of user memory on generated responses.
Expert~C initially proposed an influence budget to regulate the effects of uncertain preferences.
Expert~B challenged the reliance on confidence alone, arguing that memory influence must also account for lifecycle states, such as whether a memory is current, superseded, or context-specific.
The initial draft combined these perspectives but left the influence budget loosely defined.
In response, the experts jointly refined this mechanism: Expert~A specified four discrete influence levels, Expert~B reinforced their lifecycle constraints, and Expert~C proposed applying these interventions to a shared candidate pool before generation.
The revised draft incorporated these contributions, turning the qualitative influence budget into explicit and testable operations.

%% file: sections/conclusion.tex
\section{Conclusion}
\label{sec:conclusion}

We introduced \method, a framework for research idea generation through multi-agent debate.
Its key feature is constructing agent expertise from insights distilled across related papers, giving each agent a distinct research perspective grounded in supporting findings.
Evaluations on 100 NLP research topics show that \method achieves the highest average Elo among the compared methods under both LLM judges in both online and offline knowledge settings.

Our findings suggest that how agents acquire and use scientific knowledge is an important consideration in automated research.
Literature-grounded expert debate could serve as an ideation component in larger automated research workflows, connecting prior research to collaborative idea development.

%% file: sections/limitations.tex

\section*{Limitations}
\label{sec:limitations}

Our evaluation covers only NLP research topics, leaving the applicability of \method to other scientific disciplines untested.
Moreover, since \method starts from a predefined topic and builds expertise from existing literature, its ability to independently discover emerging or entirely new research directions remains limited.

\method generates research proposals without executing the proposed experiments or incorporating experimental feedback.
Our evaluation therefore assesses the quality of research ideas before execution, rather than whether they lead to meaningful scientific findings in practice.
Evaluating ideas through experimental execution would provide stronger evidence of their research value.

Finally, our evaluation relies primarily on LLM judges.
Although we include human evaluation to assess agreement with LLM judgments, it covers only a small subset of topics.
LLM-based assessments may not fully reflect expert judgments of research quality.

%% file: sections/ethics.tex
\section*{Ethical considerations}
\label{sec:ethics}

Our framework relies on LLMs to synthesize scientific literature and generate research proposals.
Although the proposals are grounded in prior research, they may contain unsupported claims, inaccurate interpretations of existing work, or ideas that are not genuinely novel.
These proposals have not been experimentally validated and require expert review before use.
Our evaluation also relies primarily on LLM judges, which may favor particular writing styles or types of proposals.
To assess this potential bias, we report results from two LLM judges separately and compare their judgments with human rankings on a subset of topics.

Automated research idea generation could also facilitate the large-scale production of unvetted proposals or submissions.
We intend \method to support researchers in exploring and critically assessing ideas, rather than replacing human judgment or scientific validation.

\paragraph{Use of AI assistants}
We used GPT-5.6 Luna for research topic generation and proposal development, GPT-5.6 Sol and Claude Opus 4.8 for evaluation, and Claude Opus 5.5 to categorize judges' explanations.
We used ChatGPT and Claude to assist with manuscript writing and revision.
The authors reviewed the AI-assisted content and code and take full responsibility for the paper.

%% file: sections/acknowledgments.tex
\section*{Acknowledgments}

This research was developed with funding from the Defense Advanced Research Projects Agency's (DARPA) SciFy program (Agreement No. HR00112520300) and is based upon work supported in part by the Office of the Director of National Intelligence (ODNI), Intelligence Advanced Research Projects Activity (IARPA), via 56000026C0019 (the BENGAL program). The views and conclusions contained herein are those of the authors and should not be interpreted as necessarily representing the official policies, either expressed or implied, of DARPA, ODNI, IARPA, the Department of Defense, or the U.S. Government. The U.S. Government is authorized to reproduce and distribute reprints for governmental purposes notwithstanding any copyright annotation therein.

%% file: appendix/appendix.tex
\renewcommand{\topfraction}{0.9}
\renewcommand{\bottomfraction}{0.8}
\renewcommand{\textfraction}{0.07}
\renewcommand{\floatpagefraction}{0.85}
\setlength{\textfloatsep}{12pt plus 2pt minus 4pt}

\section*{Appendix}

\input{appendix/topic_selection}

\input{appendix/offline_setting}

\input{appendix/implemetation_details}

\input{appendix/baseline_config}

\input{appendix/additional_evaluation}

\input{appendix/human_annotation}

\input{appendix/generation_process_case}

\section{Prompt Templates}
\label{sec:appendix-prompts}

The prompts are organized in execution order, covering expertise construction, agent initialization, problem selection, proposal development through debate, and final synthesis.
Pairwise evaluation is presented separately.
Table~\ref{tab:prompt-directory} summarizes the prompt groups and their inputs and outputs.

We present the online configuration with three expert agents, each initialized with an insight supported by at least two papers.
Fixed configuration values are included directly in the prompts.
For related-paper filtering, we show the first expansion level with a limit of five papers per parent; the second and third levels use limits of three and two, respectively.
Square-bracketed placeholders, such as \textit{[Research topic]}, represent runtime inputs described alongside each prompt.
Each expert's insight and supporting papers are provided as a single expertise block.
Shared proposal formats are shown once and inserted in full at runtime, while tool responses are summarized by their fields and constraints.
Explanatory notes are kept separate from the actual model instructions.

For expert and moderator calls, user messages contain the available discussion history and the current instruction.
The blind problem-ranking step instead receives only a neutral system message, the topic, and an anonymous shortlist.
The critic receives the current draft and the latest discussion summary, when available, without access to the full debate history.
System messages and step-specific instructions are presented separately, with speech-length limits included in the relevant expert instructions.

\onecolumn
\input{tables/prompt_directory}

\subsection{Expertise construction}
\label{sec:prompts-expertise}
\input{tables/prompts_graph}

\subsection{Agent setup}
\label{sec:prompts-setup}
\input{tables/prompts_roles}

\subsection{Stage 1: Problem selection}
\label{sec:prompts-problem}
\input{tables/prompts_problem}

\subsection{Stage 2: Proposal development and debate}
\label{sec:prompts-debate}
\input{tables/prompts_outputs}
\input{tables/prompts_debate}

\subsection{Stage 3: Final synthesis}
\label{sec:prompts-synthesis}
\input{tables/prompts_synthesis}

\subsection{Idea evaluation}
\label{sec:prompts-evaluation}
\input{tables/prompts_evaluation}
\twocolumn

%% file: appendix/topic_selection.tex
\section{Topic Selection}
\label{sec:appendix-topic-selection}

We construct the topic pool from the ACL Anthology's 2025 main proceedings for ACL, EMNLP, NAACL, and COLING, together with the ACL, EMNLP, and NAACL Findings volumes.
We exclude demonstrations, tutorials, student research workshops, and industry tracks, and require an abstract of at least 200 characters.
This yields 8,207 eligible papers, from which we uniformly sample 600 without replacement.
Research-area labels are used to inspect coverage after sampling, without imposing area quotas.

GPT-5.6 Luna derives one topic from each sampled paper's title and abstract.
The model is instructed to express the underlying research problem in at most 12 words, without naming specific models, datasets, or methods.
Papers without a suitable research problem may be rejected, including surveys and resource releases that pose no research question.
The finalized run rejects 10 surveys and 4 resource papers, leaving 586 candidate topics.

We embed the topics with Qwen3-Embedding-0.6B and process them in the original sampling order.
A topic is removed if its cosine similarity to any previously retained topic exceeds 0.85.
This removes 30 duplicates, leaving 556 unique topics.

We randomly sample 100 topics from the deduplicated pool for evaluation and use the same topics for all methods and both knowledge settings.
In the same call that derives a topic, GPT-5.6 Luna also assigns the source paper one of the 22 ACL Rolling Review research areas.
Table~\ref{tab:topic-areas} reports the area distribution.
The 100 topics span 20 of these areas, and no area contributes more than 12 topics.

\input{tables/topic_areas}

%% file: tables/topic_areas.tex
\begin{table}[!htbp]
    \centering
    \small
    \caption{Research areas of the 100 evaluation topics. Each source paper is assigned one ACL Rolling Review area by GPT-5.6 Luna when its topic is derived.}
    \label{tab:topic-areas}
    \setlength{\tabcolsep}{4pt}
    \begin{tabular}{@{}>{\raggedright\arraybackslash}p{0.78\columnwidth}r@{}}
        \toprule
        Research area & Topics \\
        \midrule
        Resources and Evaluation & 12 \\
        Ethics, Bias, and Fairness & 10 \\
        Multimodality and Language Grounding to Vision, Robotics and Beyond & 10 \\
        Efficient/Low-Resource Methods for NLP & 9 \\
        Information Retrieval and Text Mining & 8 \\
        Question Answering & 8 \\
        Generation & 6 \\
        Machine Learning for NLP & 6 \\
        Dialogue and Interactive Systems & 5 \\
        Interpretability and Analysis of Models for NLP & 5 \\
        Multilingualism and Language Diversity & 4 \\
        NLP Applications & 4 \\
        Information Extraction & 2 \\
        Language Modeling & 2 \\
        Linguistic Theories, Cognitive Modeling, and Psycholinguistics & 2 \\
        Machine Translation & 2 \\
        Sentiment Analysis, Stylistic Analysis, and Argument Mining & 2 \\
        Computational Social Science and Cultural Analytics & 1 \\
        Phonology, Morphology, and Word Segmentation & 1 \\
        Semantics: Lexical and Sentence-Level & 1 \\
        \midrule
        Total & 100 \\
        \bottomrule
    \end{tabular}
\end{table}

%% file: appendix/offline_setting.tex
\section{Offline Knowledge Setting}
\label{sec:appendix-offline}


The offline setting replaces the Semantic Scholar API with release 1.0 of the Scientific Contribution Graph \citep{jansen-2026-scientific-contribution-graph}, a local corpus of 230,454 papers whose coverage ends in 2025.
The graph retrieval stage uses different seed selection and neighbor expansion procedures, while insight distillation, expertise formation, and debate remain unchanged.

\paragraph{From contributions to a paper graph}
The Scientific Contribution Graph represents scientific contributions as nodes and their prerequisite relationships as directed edges.
These relationships are identified from paper texts and citations through LLM-based extraction and alignment.
We map contribution nodes to their source papers to construct a paper-level graph.
Two papers are connected when a contribution in one paper depends on a contribution in the other, with edges directed from the prerequisite paper to the dependent paper.

\paragraph{Seed retrieval and expansion}
Without a search API, we find seed candidates by embedding similarity between the topic and the corpus contributions, using Qwen3-Embedding-0.6B.
The top five candidates, with recent papers ranked first, go to the same seed-selection prompt as in the online setting.
During expansion, we retrieve both incoming and outgoing neighbors from the mapped prerequisite graph, replacing the recommendation set $\mathcal{C}(v)$ used in the online setting.
We select the top 20 candidates based on contribution relevance to the topic and apply the same LLM filtering procedure as in the online setting, with additional instructions to favor recent papers and filter out unsuitable dataset or toolkit references.
If the resulting graph is too small or lacks papers from 2024 or later, we retry with the next seed candidate.

\paragraph{Comparing the two settings}
Offline graphs contain 38.0 papers on average, compared with 43.4 online, and the two settings select the same seed paper for 9 of the 100 topics.
Because the settings also differ in corpus, edge relation, and paper evidence, we do not attribute differences between their results to the knowledge source alone.
We instead treat them as two configurations of \method, each compared with the same baseline proposals, which were generated once per topic.
ResearchAgent retains the online seed specified in Section~\ref{sec:baselines}, so it usually starts from a different paper than \method in the offline comparison.

%% file: appendix/implemetation_details.tex
\section{Implementation Details}
\label{sec:appendix-implementation}

\paragraph{Graph construction}

Given an input topic, we use GPT-5.6 Luna to select a seed paper from recent Semantic Scholar search results, prioritizing topical relevance and representativeness of the research area.
Starting from the seed, we retrieve up to 20 related candidates per expanded paper and retain up to 5, 3, and 2 new papers per parent across three expansion levels.
This gives an upper bound of 51 papers, including the seed.
The graph is passed to a single call that extracts three insights, each supported by at least two papers.

\paragraph{Generation budgets}
The configured output allowances are 8000 tokens for insight extraction, 2500 per expert speech, 4000 per critic review, and 6000 for the final proposal rewrite.
For reasoning models, the client adds a 2000-token reserve to each output allowance to accommodate reasoning tokens.

\input{tables/cost_stages}

\paragraph{Cost and efficiency}
Table~\ref{tab:cost-stages} breaks down the generation cost of \method in the online setting.
We estimate cost from logged token counts at GPT-5.6 Luna prices of \$0.20 per million input tokens, \$0.02 per million cached input tokens, and \$1.20 per million output tokens.
Generating one proposal takes 42.5 model calls and \$0.104 on average.
Summed over stages, one topic takes 7.1 minutes on average.

\input{tables/cost_methods}

Table~\ref{tab:cost-methods} compares generation cost across methods using each system's own usage log.
\method costs about as much as ResearchAgent, 1.6 times as much as Chain of Ideas, and more than 30 times as much as the idea generation stage of AI Scientist-v2.
Figure~\ref{fig:results-radar} shows that cost alone does not explain the ranking: AI Scientist-v2 has the second-highest average Elo despite its lower cost, while ResearchAgent ranks third despite costing slightly more than \method.

%% file: tables/cost_stages.tex
\begin{table}[!htbp]
    \centering
    \small
    \caption{Generation cost of \method per topic by stage, averaged over the 100 online topics. Cost is in US cents. Time sums the recorded stage durations and includes Semantic Scholar API latency. Share is the stage's percentage of total cost; Proposal dev.\ denotes proposal development.}
    \label{tab:cost-stages}
    \setlength{\tabcolsep}{2pt}
    \begin{tabular*}{\columnwidth}{@{\extracolsep{\fill}}lrrrr@{}}
        \toprule
        Stage & Calls & Cost (\textcent) & Time (s) & Share \\
        \midrule
        Seed selection        &  0.9 & 0.04 &   6.0 &  0.4 \\
        Graph construction    & 18.6 & 2.61 & 111.0 & 25.2 \\
        Insight extraction    &  1.0 & 0.83 &  32.4 &  8.0 \\
        \midrule
        Problem selection     &  8.0 & 0.92 &  57.4 &  8.8 \\
        Proposal dev.\        & 13.0 & 5.27 & 202.4 & 50.9 \\
        Final synthesis       &  1.0 & 0.68 &  18.6 &  6.6 \\
        \midrule
        Total                 & 42.5 & 10.36 & 427.7 & 100.0 \\
        \bottomrule
    \end{tabular*}
\end{table}

%% file: tables/cost_methods.tex
\begin{table}[!htbp]
    \centering
    \small
    \caption{Generation cost per topic in the online setting, averaged over 100 topics and taken from each system's own usage log. All methods use GPT-5.6 Luna at the same token prices. Tokens are in thousands; input includes cached input where logged. AI Scientist-v2 runs only its idea generation stage; Chain of Ideas includes embedding calls.}
    \label{tab:cost-methods}
    \setlength{\tabcolsep}{4pt}
    \begin{tabular*}{\columnwidth}{@{\extracolsep{\fill}}lrrr@{}}
        \toprule
        Method & Cost (\textcent) & Input (k) & Output (k) \\
        \midrule
        AI Scientist-v2 &  0.32 &   6.7 &  1.8 \\
        Chain of Ideas  &  6.48 & 121.7 & 34.4 \\
        ResearchAgent   & 10.88 & 237.3 & 57.0 \\
        \method         & 10.36 & 321.1 & 35.9 \\
        \bottomrule
    \end{tabular*}
\end{table}

%% file: appendix/baseline_config.tex
\section{Baseline Configurations}
\label{sec:appendix-baselines}


\input{tables/baseline_config}

We run each baseline from its official code using GPT-5.6 Luna, retaining its stages, iteration counts, retrieval, and selection rules.
Table~\ref{tab:baseline-config} summarizes the configuration of each baseline and how its runs executed.
We adapt API clients, output parsing, and usage logging, with the generation-setting and input changes described below.
Three settings differ from the original releases.
First, GPT-5.6 Luna does not accept a custom sampling temperature, so all baselines use the API default.
Second, AI Scientist-v2 and Chain of Ideas keep their default output limits of 4,096 and 4,000 tokens per call (16,000 for Chain of Ideas' auxiliary model), whereas we raise ResearchAgent's default of 1,000 tokens to 4,096.
On a reasoning model these limits cover both reasoning and answer tokens.
Third, ResearchAgent receives the topic text in its problem prompts in addition to the seed paper, so that its problems address the same topic as the other methods.

Although the final-rewrite allowance of \method listed in Appendix~\ref{sec:appendix-implementation} is higher, observed usage is at most 2,820 output tokens (about 2,170 on average) in both knowledge settings, below the baselines' 4,000--4,096-token limits for primary generation calls.

%% file: tables/baseline_config.tex
\begin{table*}[t]
\centering
\small
\caption{Baseline configurations using official implementations and GPT-5.6 Luna. Adaptations are detailed in Appendix~\ref{sec:appendix-baselines}. Executed counts are averaged over 100 online evaluation topics, with ranges in parentheses where available.}
\label{tab:baseline-config}
\setlength{\tabcolsep}{4pt}
\renewcommand{\arraystretch}{1.1}
\begin{tabular}{@{}>{\raggedright\arraybackslash}p{0.12\textwidth}>{\raggedright\arraybackslash}p{0.17\textwidth}>{\raggedright\arraybackslash}p{0.2\textwidth}>{\raggedright\arraybackslash}p{0.28\textwidth}>{\raggedright\arraybackslash}p{0.15\textwidth}@{}}
\toprule
\textbf{Baseline} & \textbf{Input} & \textbf{Literature} & \textbf{Generation and refinement} & \textbf{Evaluated output} \\
\midrule

AI Scientist-v2 (ideation only)
& Topic used as workshop description
& Semantic Scholar search; 10 results per query; 1.7 searches (1--4)
& 1 idea; up to 5 rounds of searching or finalizing; 2.7 rounds (2--5), ending upon finalization
& Generated idea \\

Chain of Ideas
& Topic
& Semantic Scholar citations and references; 1 chain of up to 5 papers; 3.8 papers (1--5)
& Trend summarization; repeated idea generation until passing the novelty check, 1.3 rounds (1--6); experiment design with 1 refinement round
& Final idea \\

ResearchAgent
& Topic and \method-selected seed paper
& Up to 10 seed-paper references and an entity store from the original release; 7.5 references
& Problem, method, and experiment design; each generated and refined twice using feedback from 5 reviewing agents
& Highest-rated problem, method, and experiment design \\

\bottomrule
\end{tabular}
\end{table*}

%% file: appendix/additional_evaluation.tex

\section{Additional Evaluation Results}
\label{sec:appendix-results}

This section supplements the main results in four parts.
Section~\ref{sec:appendix-direct} reports detailed scores and compares \method directly with each baseline.
Section~\ref{sec:appendix-feasibility} examines why \method falls behind AI Scientist-v2 on feasibility.
Section~\ref{sec:appendix-rationales} analyzes the LLM judges' reasoning across all dimensions and baselines.
Section~\ref{sec:appendix-reliability} tests whether the LLM evaluation behind these results is reliable.

\subsection{Detailed results and direct comparison}
\label{sec:appendix-direct}

Table~\ref{tab:main_results} reports the Elo ratings shown in Figure~\ref{fig:results-radar} for both knowledge settings, five dimensions, and two judges.
Elo ratings summarize each method's overall performance across all pairwise comparisons.
To show how \method performs against each baseline individually, Table~\ref{tab:direct_results} reports direct comparison scores, counting wins as 1 and ties as 0.5.

\input{tables/main_results}
\input{tables/direct_results}

The direct comparisons show similar patterns across both knowledge settings.
Against Chain of Ideas and ResearchAgent, \method scores above 68\% on every dimension under both judges, with average scores ranging from 86.8\% to 92.8\%.
Against AI Scientist-v2, \method achieves average scores of 70.8\% online and 73.8\% offline under GPT-5.6 Sol, and 61.5\% and 63.3\% under Claude Opus 4.8, with all four confidence intervals above 50\%.
Scores for novelty, significance, and expected effectiveness remain at or above 74\%, while feasibility is substantially lower (16.0--36.5\%), as further examined in Section~\ref{sec:appendix-feasibility}.

Clarity is the only dimension where the two judges favor different methods in both settings.
GPT-5.6 Sol favors \method over AI Scientist-v2 (77.0\% online; 74.5\% offline), whereas Claude Opus 4.8 favors AI Scientist-v2 (34.0\% online; 38.5\% offline for \method).
Section~\ref{sec:appendix-dimension-agreement} further examines this disagreement.

\subsection{Feasibility trade-off}
\label{sec:appendix-feasibility}

Table~\ref{tab:main_results} shows that AI Scientist-v2 outperforms \method on feasibility across both knowledge settings and judges.
To better understand this gap, we analyze their 200 online pairwise comparisons per judge, where \method loses 148 under GPT-5.6 Sol and 161 under Claude Opus 4.8.

\input{tables/feasibility_mentions}

\paragraph{Judge rationale analysis}
We examine the rationales favoring AI Scientist-v2, retaining 127 from GPT-5.6 Sol and 143 from Claude Opus 4.8 that explicitly reference both proposals by their anonymous labels.
Table~\ref{tab:feasibility-mentions} summarizes feasibility-related factors identified through keyword matching.
For \method, the rationales frequently mention human annotation and multiple components, while Claude Opus 4.8 also highlights validation requirements such as preregistration and audits.
Compute-related concerns appear in fewer than 10\% of the rationales.
For AI Scientist-v2, more than 80\% mention simple or standard resources.
Only one of the 270 rationales explicitly describes \method as infeasible, impossible, or impractical.

\paragraph{Discussion}
These patterns suggest that the feasibility gap may be associated more with the perceived scope and validation requirements of the proposed studies than with computational demands.
However, keyword matches indicate which factors appear in the rationales, rather than which factors determine the judges' preferences.

\input{appendix/rationale_analysis}

\subsection{Reliability of the LLM evaluation}
\label{sec:appendix-reliability}

We conduct four analyses to assess the consistency and potential biases of our LLM-based evaluation.
Section~\ref{sec:appendix-judge-consistency} examines whether judgments remain consistent when presentation order is reversed and how often the two judges agree.
Section~\ref{sec:appendix-human-evaluation} compares LLM judgments with human annotations.
Section~\ref{sec:appendix-dimension-agreement} examines how agreement varies across evaluation dimensions.
Section~\ref{sec:appendix-length} investigates whether proposal length is associated with judge preferences.

\subsubsection{Consistency within and across LLM judges}
\label{sec:appendix-judge-consistency}

\input{tables/evaluation_reliability}

Table~\ref{tab:evaluation-reliability} reports consistency within each LLM judge under reversed presentation orders and agreement between the two judges.
For within-judge consistency, decisions remain unchanged in at least 89.0\% of comparisons for GPT-5.6 Sol and 80.3\% for Claude Opus 4.8 on novelty, feasibility, and expected effectiveness across both settings.
Claude Opus 4.8 shows lower consistency on significance and clarity (72.3--76.0\%).
For cross-judge agreement, the two judges agree on at least 83.1\% of comparisons on novelty, feasibility, and expected effectiveness, but only 68.9--72.5\% on significance and clarity.
Overall, judgments are more consistent on novelty, feasibility, and expected effectiveness, with greater variation on significance and clarity.

\input{appendix/human_evaluation}

\subsubsection{Proposal length}
\label{sec:appendix-length}

\input{tables/length_analysis}

The evaluation prompt instructs judges to assess proposal quality regardless of length (Section~\ref{sec:evaluation}).
We examine whether proposal length is associated with judge preferences.
Proposal lengths vary substantially across methods: AI Scientist-v2 averages 931 words, \method 1,297 online and 1,281 offline, Chain of Ideas 2,175, and ResearchAgent 3,274.
Table~\ref{tab:length-analysis} shows that longer proposals win only 18.1--46.5\% of non-tied comparisons across dimensions, settings, and judges, considering all six method pairs.

We also examine \method against AI Scientist-v2, its closest competitor, which outperforms \method on some evaluation dimensions.
Across topics, the correlation between proposal length differences and \method's average score is weak ($r=0.06$--$0.21$), with dimension-level correlations ranging from $-0.08$ to $0.27$.
Overall, these results provide no clear evidence of a systematic preference for longer proposals.

%% file: tables/main_results.tex
\begin{table*}[!tp]
\centering
\small
\caption{Pairwise Elo ratings ($\uparrow$). Setup lists the knowledge setting, topic count $n$, and judge. Eff. denotes expected effectiveness; Avg. is the mean of the five dimensions. Shading highlights \method; bold numbers mark the highest rating within each setup.}
\label{tab:main_results}
\setlength{\tabcolsep}{3pt}
\renewcommand{\arraystretch}{1.12}
\begin{tabular*}{\textwidth}{@{\extracolsep{\fill}}llrrrrrr@{}}
\toprule
{\scriptsize Setup} & Method & Novelty & Significance & Feasibility & Clarity & Eff. & Avg. \\
\midrule
\multirow{4}{*}{\scriptsize\shortstack[l]{Online ($n=100$)\\GPT-5.6 Sol}} & AI Scientist-v2 & 1021.0 & 976.7 & \textbf{1248.1} & 1111.0 & 1001.4 & 1071.6 \\
 & Chain of Ideas & 789.3 & 850.7 & 855.8 & 692.2 & 576.2 & 752.8 \\
 & ResearchAgent & 775.9 & 983.9 & 791.3 & 861.3 & 1027.7 & 888.0 \\
 & \cellcolor{orange!15}\textbf{\method (ours)} & \textbf{1413.8} & \textbf{1188.6} & 1104.8 & \textbf{1335.5} & \textbf{1394.6} & \textbf{1287.5} \\
\midrule
\multirow{4}{*}{\scriptsize\shortstack[l]{Online ($n=100$)\\Claude Opus 4.8}} & AI Scientist-v2 & 1027.5 & 974.2 & \textbf{1267.2} & \textbf{1258.5} & 1080.7 & 1121.6 \\
 & Chain of Ideas & 701.0 & 789.9 & 882.5 & 741.8 & 556.4 & 734.3 \\
 & ResearchAgent & 749.6 & 975.2 & 844.2 & 871.1 & 970.2 & 882.1 \\
 & \cellcolor{orange!15}\textbf{\method (ours)} & \textbf{1521.9} & \textbf{1260.7} & 1006.1 & 1128.6 & \textbf{1392.7} & \textbf{1262.0} \\
\midrule
\multirow{4}{*}{\scriptsize\shortstack[l]{Offline ($n=100$)\\GPT-5.6 Sol}} & AI Scientist-v2 & 1030.3 & 952.4 & \textbf{1210.4} & 1147.1 & 980.6 & 1064.2 \\
 & Chain of Ideas & 797.9 & 847.8 & 866.4 & 651.2 & 598.3 & 752.3 \\
 & ResearchAgent & 808.2 & 998.0 & 785.0 & 830.6 & 1010.0 & 886.4 \\
 & \cellcolor{orange!15}\textbf{\method (ours)} & \textbf{1363.5} & \textbf{1201.8} & 1138.2 & \textbf{1371.2} & \textbf{1411.1} & \textbf{1297.2} \\
\midrule
\multirow{4}{*}{\scriptsize\shortstack[l]{Offline ($n=100$)\\Claude Opus 4.8}} & AI Scientist-v2 & 1055.7 & 972.0 & \textbf{1240.8} & \textbf{1252.3} & 1051.4 & 1114.4 \\
 & Chain of Ideas & 735.3 & 797.3 & 874.1 & 716.7 & 547.2 & 734.1 \\
 & ResearchAgent & 786.3 & 983.5 & 835.2 & 865.2 & 968.9 & 887.8 \\
 & \cellcolor{orange!15}\textbf{\method (ours)} & \textbf{1422.7} & \textbf{1247.2} & 1049.8 & 1165.9 & \textbf{1432.5} & \textbf{1263.6} \\
\bottomrule
\end{tabular*}
\end{table*}

%% file: tables/direct_results.tex
\begin{table*}[t]
\centering\small
\caption{Direct comparison scores (\%, $\uparrow$) of \method against each baseline, with half credit for ties. Each dimension uses 200 online or 200 offline judgments. A score above 50 means \method is preferred more often; Eff. denotes expected effectiveness and Avg. equally weights all five dimensions. Brackets give 95\% percentile bootstrap intervals for Avg., resampling topics 10,000 times while retaining both orders and all dimensions together.}
\label{tab:direct_results}
\setlength{\tabcolsep}{5pt}
\begin{tabular*}{\textwidth}{@{\extracolsep{\fill}}lrrrrrl@{}}
\toprule
Opponent & Novelty & Significance & Feasibility & Clarity & Eff. & Avg. [95\% CI] \\
\midrule
\multicolumn{7}{l}{\textit{Online, GPT-5.6 Sol}} \\
AI Scientist-v2 & 89.0 & 74.0 & 26.0 & 77.0 & 88.0 & 70.8 [67.6, 73.8] \\
Chain of Ideas & 97.5 & 89.0 & 80.0 & 97.5 & 100.0 & 92.8 [90.7, 94.8] \\
ResearchAgent & 98.5 & 75.5 & 90.0 & 94.0 & 89.5 & 89.5 [87.0, 91.9] \\
\midrule
\multicolumn{7}{l}{\textit{Online, Claude Opus 4.8}} \\
AI Scientist-v2 & 93.0 & 80.8 & 16.0 & 34.0 & 83.8 & 61.5 [58.0, 64.9] \\
Chain of Ideas & 99.5 & 95.0 & 68.5 & 86.2 & 100.0 & 89.8 [87.8, 91.8] \\
ResearchAgent & 100.0 & 84.2 & 73.2 & 84.5 & 92.2 & 86.8 [84.0, 89.6] \\
\midrule
\multicolumn{7}{l}{\textit{Offline, GPT-5.6 Sol}} \\
AI Scientist-v2 & 86.5 & 82.0 & 36.5 & 74.5 & 89.5 & 73.8 [70.5, 77.1] \\
Chain of Ideas & 96.0 & 86.5 & 81.0 & 98.5 & 99.5 & 92.3 [90.3, 94.2] \\
ResearchAgent & 96.0 & 72.5 & 90.5 & 98.5 & 92.5 & 90.0 [87.8, 92.2] \\
\midrule
\multicolumn{7}{l}{\textit{Offline, Claude Opus 4.8}} \\
AI Scientist-v2 & 87.8 & 78.8 & 23.0 & 38.5 & 88.5 & 63.3 [60.0, 66.5] \\
Chain of Ideas & 98.5 & 93.2 & 72.0 & 90.8 & 99.2 & 90.8 [88.7, 92.8] \\
ResearchAgent & 97.5 & 83.2 & 79.8 & 86.5 & 94.8 & 88.3 [85.9, 90.7] \\
\bottomrule
\end{tabular*}
\end{table*}

%% file: tables/feasibility_mentions.tex
\begin{table}[!htbp]
    \centering
    \small
    \caption{What feasibility rationales mention when AI Scientist-v2 wins (\% of rationales). $n$ counts rationales that refer to both proposals by their anonymous labels. Categories are keyword matches on the text describing each proposal, so a mention is not a criticism; a rationale can match several categories.}
    \label{tab:feasibility-mentions}
    \setlength{\tabcolsep}{4pt}
    \begin{tabular*}{\columnwidth}{@{\extracolsep{\fill}}lrr@{}}
        \toprule
         & Sol & Opus \\
         & {\scriptsize $n=127$} & {\scriptsize $n=143$} \\
        \midrule
        \multicolumn{3}{l}{\textit{Text describing \textup{\method} mentions}} \\
        Human annotation or studies     & 44.9 & 49.0 \\
        Multiple components             & 38.6 & 49.7 \\
        Validation protocol             & 28.3 & 48.3 \\
        New or counterfactual data      & 11.0 & 20.3 \\
        Compute or training scale       &  9.4 &  5.6 \\
        Any of the above                & 80.3 & 88.1 \\
        \midrule
        \multicolumn{3}{l}{\textit{Text describing AI Scientist-v2 mentions}} \\
        Simple or standard resources    & 80.3 & 83.9 \\
        \bottomrule
    \end{tabular*}
\end{table}

%% file: appendix/rationale_analysis.tex

\subsection{LLM Judge Reasoning Analysis}
\label{sec:appendix-rationales}

For every pairwise judgment, each judge writes a one- or two-sentence explanation per evaluation dimension.
We analyze all 6,000 online explanations that compare \method with a baseline: 300 topic--baseline pairs, two presentation orders, five dimensions, and two judges.
We derived the properties in Table~\ref{tab:rationale-taxonomy} through LLM-assisted open coding of a stratified sample of 170 explanations, oversampling comparisons where \method loses, the judges disagree, or the verdict changes with presentation order.
The authors reviewed and revised the resulting categories, and the definitions were refined until two independent LLM coders applied them consistently (Cohen's $\kappa \geq 0.90$ per property).
The labeler also codes a readability property, which we do not report because it overlaps with the clarity dimension.

\input{tables/rationale_taxonomy}

Claude Opus 5.5 labels each explanation through the Batch API with a JSON schema that fixes the output to one label per property.
The labeler sees the evaluated dimension, the winner, and the explanation exactly as the judge wrote it, with the proposals named Idea0 and Idea1 by presentation order.
When an explanation cites an execution barrier, the labeler also records its type, using the four types in Table~\ref{tab:rationale-taxonomy}.

\begin{figure*}[!t]
    \centering
    \includegraphics[width=\textwidth]{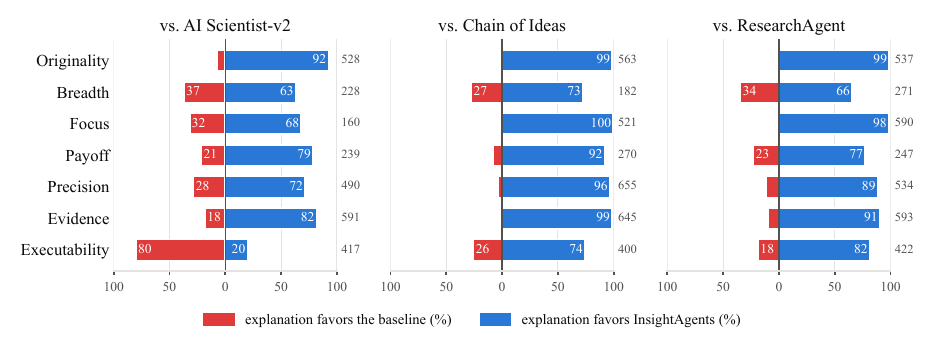}
    \caption{LLM judges' preferences by proposal property, based on their stated reasoning. Bars show the percentage of explanations favoring \method (right) or the baseline (left), among those discussing each property. Numbers beside the bars give the total count of explanations discussing that property. Results pool both judges in the online setting.}
    \label{fig:rationale-categories}

    \vspace{1.2em}
    \includegraphics[width=\textwidth]{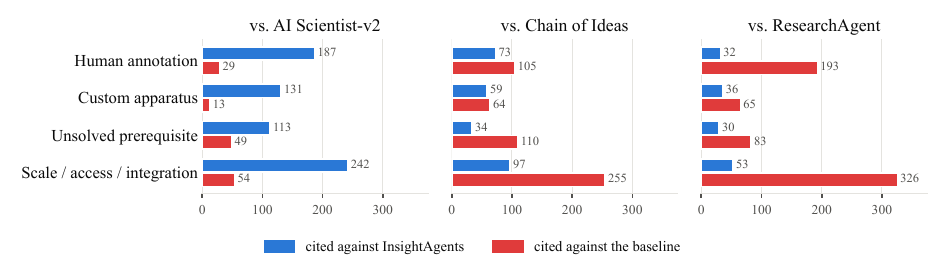}
    \caption{Execution barriers identified in the LLM judges' stated reasoning. Bars show the number of explanations citing each barrier against \method or the baseline. Each explanation may cite multiple barrier types. Results pool both judges in the online setting.}
    \label{fig:rationale-burdens}
\end{figure*}

To test whether the labeler favors our system when it knows which one it is, we repeated the labeling with a prompt that identified our proposal.
The two runs agree on 97.7\% of property labels, and no share in Figure~\ref{fig:rationale-categories} changes by more than 3.0 points.

Figure~\ref{fig:rationale-burdens} breaks down the execution barriers. Against AI Scientist-v2, barriers are cited against \method far more often than against the baseline, with the relative largest gaps for human annotation (187 vs.\ 29) and custom apparatus (131 vs.\ 13). Among the 309 feasibility judgments that \method loses to AI Scientist-v2, every explanation cites at least one barrier against \method: scale, data access, or integration in 76.1\%, human annotation in 52.8\%, custom apparatus in 39.8\%, and an unsolved prerequisite in 34.6\%. These shares are consistent with the keyword analysis in Table~\ref{tab:feasibility-mentions}. Against Chain of Ideas and ResearchAgent the direction reverses, and every barrier type is cited more often against the baseline.

%% file: tables/rationale_taxonomy.tex
\begin{table*}[!htbp]
    \centering
    \small
    \caption{Proposal properties used to label judge explanations. For each explanation and property, the labeler records which proposal the judge favors, or that the property is not discussed. A property can be cited under any evaluation dimension.}
    \label{tab:rationale-taxonomy}
    \renewcommand{\arraystretch}{1.15}
    \begin{tabularx}{\textwidth}{@{}l>{\raggedright\arraybackslash}X>{\raggedright\arraybackslash}X@{}}
        \toprule
        Property & Favors a proposal when the judge says it & Disfavors a proposal when the judge says it\\
        \midrule
        Originality & reframes the problem, the unit of analysis, or the object of measurement, or introduces a distinctive mechanism & assembles familiar components, follows an established experimental paradigm, or is incremental to existing work \\
        Breadth & would generalize across models, tasks, domains, languages, or architectures & is restricted to one dataset, language pair, model family, or domain \\
        Focus & centers on one decisive question whose outcome is informative either way & is diffuse, spread across many goals or loosely coupled components \\
        Payoff & delivers a kind of result the judge values, such as a usable method or a validity check on existing evaluation & delivers a kind of result the judge discounts, such as one that is merely diagnostic \\
        Precision & defines estimands, formulas, thresholds, and success or refutation criteria that tie the claim to its test & leaves the central mechanism, scoring rule, or construct undefined, or has parts that do not fit together \\
        Evidence & uses matched baselines, counterfactual pairs, oracle conditions, or controls that rule out alternative explanations & relies on circular evaluation, confounded attribution, few units, or a measure that may not capture the construct \\
        Executability & relies on standard models, public data, conventional pipelines, and a bounded primary study & requires human or expert annotation, custom apparatus, an unsolved prerequisite, or large scale, restricted data, or many subsystems \\
        \bottomrule
    \end{tabularx}
\end{table*}

%% file: tables/evaluation_reliability.tex
\begin{table}[!htbp]
    \centering
    \small
    \caption{Evaluation agreement (\%) on 100 topics per setting. Order agreement measures whether swapping presentation order preserves the winner or tie ($n=600$ pairs per dimension and setting). Judge agreement compares the two judges on identical pairs and orders ($n=1{,}200$).}
    \label{tab:evaluation-reliability}
    \setlength{\tabcolsep}{3pt}
    \begin{tabular*}{\columnwidth}{@{\extracolsep{\fill}}lrrr@{}}
        \toprule
        \multirow{2}{*}{Dimension} & \multicolumn{2}{c}{Order agreement} & \multirow{2}{*}{\shortstack{Judge\\agreement}} \\
        \cmidrule(lr){2-3}
         & Sol & Opus & \\
        \midrule
        \multicolumn{4}{l}{\textit{Online}} \\
        Novelty            & 89.5 & 88.2 & 86.8 \\
        Significance       & 83.2 & 73.3 & 71.2 \\
        Feasibility        & 92.2 & 82.3 & 83.4 \\
        Clarity            & 91.2 & 74.2 & 68.9 \\
        Exp. effectiveness & 90.5 & 86.0 & 83.1 \\
        \midrule
        \multicolumn{4}{l}{\textit{Offline}} \\
        Novelty            & 89.0 & 87.2 & 85.6 \\
        Significance       & 85.0 & 72.3 & 69.3 \\
        Feasibility        & 91.7 & 80.3 & 83.7 \\
        Clarity            & 92.0 & 76.0 & 72.5 \\
        Exp. effectiveness & 90.7 & 87.7 & 84.6 \\
        \bottomrule
    \end{tabular*}
\end{table}

%% file: appendix/human_evaluation.tex
\subsubsection{Agreement with human annotators}
\label{sec:appendix-human-evaluation}

We compare human and LLM preferences on the 150 \method--baseline instances described in Appendix~\ref{sec:appendix-human-annotation}.
For each LLM judge, we combine judgments from both presentation orders, treating equal win counts as ties.
We compute exact agreement and non-contradiction as defined in Section~\ref{sec:results}.
We report 95\% confidence intervals (in brackets) from 2,000 topic-level bootstrap resamples, preserving all comparisons within each sampled topic.
Figure~\ref{fig:human-agreement-dimensions} presents these estimates for each dimension and overall.

Figure~\ref{fig:human-agreement} shows that human and LLM judges rarely express opposing preferences.
Human--LLM non-contradiction reaches 90.0\% [87.0, 92.3] for GPT-5.6 Sol and 92.7\% [89.7, 95.3] for Claude Opus 4.8, compared with 90.0\% [86.0, 94.0] between human annotators and 93.3\% [89.3, 96.7] between LLM judges.
All four confidence intervals overlap, suggesting no clear difference in non-contradiction rates.

Exact agreement is lower: 63.7\% [55.7, 70.0] and 65.0\% [55.3, 74.3] for human--LLM comparisons, compared with 44.0\% [30.0, 58.0] between humans and 82.0\% [78.7, 84.7] between LLM judges.
The non-overlapping human--human and LLM--LLM intervals indicate substantially higher exact agreement between the two LLM judges in this sample.
Together, these results suggest that many human--LLM disagreements involve ties rather than directly opposing preferences.

\subsubsection{Agreement by dimension}
\label{sec:appendix-dimension-agreement}

\begin{figure*}[!t]
    \centering
    \includegraphics[width=\textwidth]{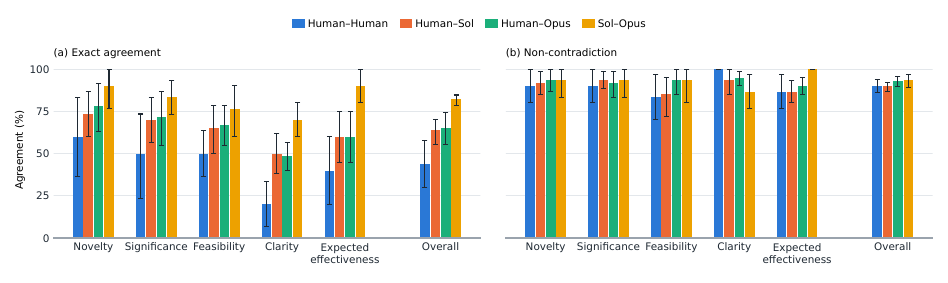}
    \caption{Agreement by dimension on the same \method--baseline comparisons as Figure~\ref{fig:human-agreement}. Each dimension contains 30 topic--pair instances, and Overall pools all 150. Error bars give 95\% topic-bootstrap intervals. Sol and Opus denote GPT-5.6 Sol and Claude Opus 4.8.}
    \label{fig:human-agreement-dimensions}
\end{figure*}

\begin{figure*}[!t]
    \centering
    \includegraphics[width=\textwidth]{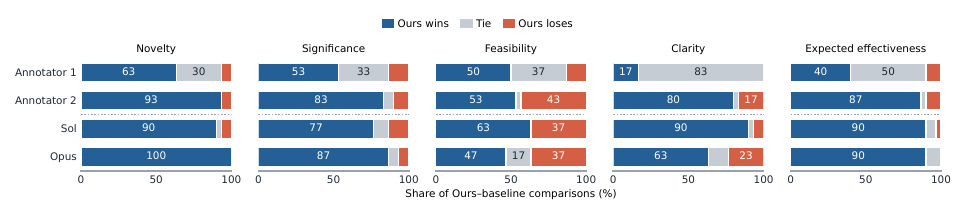}
    \caption{Outcomes of \method--baseline comparisons for each judge and dimension. Each bar gives the share of 30 topic--pair instances in which the judge prefers \method, returns a tie, or prefers the baseline. Labels show shares of at least 15\%. Ours denotes \method.}
    \label{fig:human-win-tie-loss}
\end{figure*}

Figure~\ref{fig:human-agreement-dimensions} shows that human--LLM exact agreement is highest for novelty (73.3--78.3\%) and lowest for clarity (48.3--50.0\%).
Non-contradiction remains at least 83.3\% across all judge pairs and dimensions.
However, each dimension contains only 30 topic--pair instances, resulting in wide confidence intervals.

Figure~\ref{fig:human-win-tie-loss} helps explain these differences.
Annotator~1 returns ties in 30--83\% of comparisons across dimensions, while the other judges do so in at most 17\%.
This contrast is particularly pronounced for clarity, where human--human exact agreement is only 20.0\%, despite 100.0\% non-contradiction.
Thus, low exact agreement on clarity largely reflects differences in tie frequency, although the LLM judges also show different clarity preferences in the full evaluation (Section~\ref{sec:appendix-direct}).

Feasibility shows a different pattern.
All four judges favor the baseline over \method in 13--43\% of feasibility comparisons, compared with at most 7\% for novelty.
Feasibility also has the lowest non-contradiction between human annotators (83.3\%) and between humans and GPT-5.6 Sol (85.0\%), indicating more directly opposing preferences.
This is consistent with the feasibility trade-off observed in the main evaluation.

%% file: tables/length_analysis.tex
\begin{table}[!htbp]
    \centering
    \small
    \caption{Proposal length and judgments. Top: share (\%) of decided judgments won by the longer proposal, pooled over all six method pairs and both orders. Bottom: Pearson $r$ over 100 topics between the word-count difference (\method minus AI Scientist-v2) and \method's average score across the five dimensions.}
    \label{tab:length-analysis}
    \setlength{\tabcolsep}{4pt}
    \begin{tabular*}{\columnwidth}{@{\extracolsep{\fill}}lrrrr@{}}
        \toprule
        & \multicolumn{2}{c}{Online} & \multicolumn{2}{c}{Offline} \\
        \cmidrule(lr){2-3}\cmidrule(lr){4-5}
        & Sol & Opus & Sol & Opus \\
        \midrule
        \multicolumn{5}{l}{\textit{Longer proposal wins (\%)}} \\
        Novelty            & 30.4 & 29.7 & 31.4 & 29.5 \\
        Significance       & 43.8 & 41.5 & 46.5 & 41.3 \\
        Feasibility        & 21.1 & 22.5 & 21.8 & 22.3 \\
        Clarity            & 31.4 & 18.1 & 29.2 & 18.6 \\
        Exp. effectiveness & 42.4 & 36.7 & 42.7 & 38.0 \\
        \midrule
        \multicolumn{5}{l}{\textit{\textup{\method} vs.\ AI Scientist-v2}} \\
        Length--score $r$  & 0.21 & 0.18 & 0.10 & 0.06 \\
        \bottomrule
    \end{tabular*}
\end{table}

%% file: appendix/human_annotation.tex
\section{Human Annotation Protocol}
\label{sec:appendix-human-annotation}

\input{tables/human_annotation_instructions}

We conduct a human evaluation to assess agreement with LLM judges.
Two NLP researchers independently evaluate proposals on 10 online topics selected based on their familiarity with the topics.
Each topic includes four proposals generated by \method, AI Scientist-v2, Chain of Ideas, and ResearchAgent.
We use the same proposal texts as in the LLM evaluation, without modification.

\paragraph{Annotation procedure}
Annotators follow the same evaluation criteria and judging instructions as the LLM judges, including guidance on avoiding position and length bias (Table~\ref{tab:human-annotation-instructions}).
A web-based interface presents the four proposals side by side, with method identities hidden and display positions independently randomized for each topic and annotator.
For each of the five dimensions, annotators rank all four proposals from best to worst, allowing ties (e.g., \texttt{B > A = C > D}).
Unlike LLM judges, who compare proposals pairwise, human annotators rank all four proposals together.

\paragraph{Agreement computation}
We convert each ranking into six pairwise outcomes, treating equally ranked proposals as ties.
Human Elo ratings are computed using the same procedure as the LLM evaluation (Figure~\ref{fig:human-radar}).
For agreement analysis, we consider the three \method--baseline pairs on topics completed by both annotators, yielding 150 topic--dimension--pair instances (10 topics $\times$ 5 dimensions $\times$ 3 pairs).
These outcomes are compared with LLM judgments in Section~\ref{sec:appendix-human-evaluation}.

%% file: tables/human_annotation_instructions.tex
\begin{table}[!htbp]
    \centering
    \small
    \caption{Instructions shown to human annotators. The bracketed block stands for the criteria text, which is identical to the LLM judge prompt in Table~\ref{tab:prompts-evaluation} and is not repeated here. Instructions about saving are omitted.}
    \label{tab:human-annotation-instructions}
    \begin{tabular}{@{}>{\raggedright\arraybackslash}p{0.97\columnwidth}@{}}
        \toprule
        \textbf{1. What you are asked to do}\\
        Each of the 10 topics in this study shows 4 research ideas side by side, labelled A, B, C, D. Which system wrote which idea is hidden from you, and the order is randomised per annotator.\\[3pt]
        Read them, then rank them from best to worst separately for each of the five criteria: one text box per criterion, \texttt{>} = strictly better, \texttt{=} = equally good, every label used exactly once. For example:\\
        \hspace*{1em}\texttt{B > A > C} \quad B best, then A, then C\\
        \hspace*{1em}\texttt{B > A = C} \quad B best, A and C equally good\\
        \hspace*{1em}\texttt{A = B = C} \quad all indistinguishable on this criterion\\[3pt]
        A topic counts as done only once all five criteria are ranked.\\
        \midrule
        \textbf{2. What to judge on}\\
        \textit{[The five evaluation criteria and the general rules for all criteria, copied word for word from the LLM judge prompt (Table~\ref{tab:prompts-evaluation}).]}\\[3pt]
        The LLM judge is given the criteria above word-for-word, plus this instruction, which applies to you as well:\\[3pt]
        \textit{[The note on position and length bias, copied word for word from the same prompt.]}\\
        \bottomrule
    \end{tabular}
\end{table}

%% file: appendix/generation_process_case.tex
\onecolumn

\section{Detailed Generation Case Study}
\label{sec:appendix-generation-case}

This appendix presents the online case analyzed in Section~\ref{sec:generation-process-analysis}, for the topic \textit{Personalized dialogue generation through evolving user-specific memory}.
Tables headed \textit{Generated output} reproduce saved model outputs verbatim, with shaded cells marking text written by the system; unshaded rows, step labels, and the prose between tables are written by the authors.
Paper labels such as [P18] are local to this case; each insight table ends with a key that identifies the cited papers.

\subsection{Insights and supporting findings}
\label{sec:case-insights}

Tables~\ref{tab:case-insight-a}--\ref{tab:case-insight-c} show the three insights distilled from the paper graph, one per expert, with the supporting findings each expert received.

\input{tables/case_insights}

\subsection{Final research proposal}
\label{sec:case-proposal}

Table~\ref{tab:case-proposal} shows the complete reader-facing proposal.
Its claims and experimental requirements are retained as generated.

\input{tables/case_proposal}

\subsection{Debate excerpts and proposal revisions}
\label{sec:case-revisions}

The tables below quote exact spans from the debate record; the text before each table summarizes the process.
Turn numbers refer to proposal development (Stage~2).
Two intermediate drafts were saved, after turns 1 and 2; the convergence speeches in turn 3 were incorporated during final synthesis (Stage~3).

\paragraph{Expertise informing the initial design}
Table~\ref{tab:case-initial} illustrates how the experts contributed to the initial proposal.
Expert~A emphasized comparing responses with and without memory to measure its influence, while Expert~C proposed limiting the influence of uncertain preferences.
Expert~B challenged these proposals, arguing that memory influence must also account for its lifecycle state.
Although Experts~A and C both discussed counterfactual comparisons, their proposals emphasized different aspects of the problem.

\input{tables/case_initial}

\paragraph{Making the influence budget operational}
Table~\ref{tab:case-chain-budget} shows how the influence budget evolved through debate.
The initial draft combined evidence-state reconstruction, influence control, and response comparison, but defined the budget only through qualitative examples.
The critic called for explicit, testable interventions.
In response, Expert~A specified four discrete influence levels, Expert~B reinforced their constraints, and Expert~C proposed applying the interventions to a shared structured representation before response generation.
The second draft incorporated these changes.
Following another critic review, the experts further specified matched candidate pools and deterministic decoding, restricting the generator to the post-intervention representation.
These refinements were retained in the final proposal, making the influence control mechanism more explicit and testable.

\input{tables/case_chain_budget}

\paragraph{Aligning the claim with the evaluation scope}
Table~\ref{tab:case-chain-decay} shows how the experts refined the proposal's scope through debate.
The initial draft included a mechanism to reduce the influence of uncertain memories through repeated use, aiming to prevent self-reinforcing personalization.
The critic questioned whether the proposed case-based benchmark could adequately evaluate this longitudinal mechanism.
Experts~A and B suggested deferring it, and Expert~C, who originally proposed it, agreed.
The revised draft removed the mechanism from the primary study, and the final proposal retained it only as future work.

\input{tables/case_chain_decay}

\twocolumn

%% file: tables/case_insights.tex
\begingroup\small
\setlength{\tabcolsep}{6pt}
\renewcommand{\arraystretch}{1.15}
\begin{longtable}{>{\raggedright\arraybackslash}p{\dimexpr\textwidth-2\tabcolsep\relax}}
\toprule
\textbf{Generated output:} Expert A insight --- Memory Utilization Evaluation\\
\midrule\endfirsthead
\toprule
\textbf{Generated output:} Expert A insight --- Memory Utilization Evaluation (continued)\\
\midrule\endhead
\midrule\hfill Continued on next page\\\endfoot
\bottomrule\noalign{\vskip 6pt}
\caption{Insight distilled for Expert~A (Memory Utilization Evaluation), reproduced verbatim. Shaded rows are model output; the paper key below them is added by the authors.}\label{tab:case-insight-a}\\
\endlastfoot
\rowcolor{gray!12}\textbf{Core insight.} This line keeps finding that retrieving a remembered fact is easier than recognizing when it matters and integrating it naturally into an answer. Evaluation has shifted from direct question answering toward implicit, composed, guidance, and evidence-synthesis tasks. It assumes useful memory must be judged in context, not as isolated retrieval. The field remains stuck at the boundary between recovered evidence and coherent, appropriately personalized reasoning.\\
\rowcolor{gray!12}\textbf{Historical tension.} Broader, more realistic evaluation exposes genuine failures but sacrifices the clean attribution and comparability of pointwise recall benchmarks.\\
\rowcolor{gray!12}\textbf{Grounding summary.} This reading rests on P18, P37, P44, and P5, representing the graph's evaluation and benchmarking line. The graph shows a clear progression from elicited recall toward natural integration, distributed evidence use, and decision-support evaluation.\\
\rowcolor{gray!12}\textbf{Supporting findings.}\\
\rowcolor{gray!12}\hspace*{1em}\textbf{[P18]} In a four-month deployment with 40 users, 1,872 sessions, and seven memory conditions, Direct QA ranged from 19.7\% to 70.1\% without changing satisfaction; only 7.9\% of recalled facts were naturally referenced.\\
\rowcolor{gray!12}\hspace*{1em}\textbf{[P37]} On LOCOMO-CONV, conversational framing exposed retrieval gaps for implicit and composed queries; multi-facet rewriting helped raw-turn memory but not abstractive memory.\\
\rowcolor{gray!12}\hspace*{1em}\textbf{[P44]} UtilMem found that strong factual-memory scores did not reliably predict memory utilization: retrieval alone often failed to integrate distributed evidence or reject plausible distractors.\\
\rowcolor{gray!12}\hspace*{1em}\textbf{[P5]} PRAGMA's longitudinal guidance scenarios, with evidence annotations and evolving contexts, showed current retrieval, memory, and long-context systems struggling to use evidence for personalized guidance.\\
\midrule
\footnotesize\textit{Paper key.} \textbf{[P18]} \textit{MemUse: Moving Memory Evaluation from Direct QA to Natural Integration in Long-Term Human-AI Conversation}~\citep{sumida-etal-2026-memuse}; \textbf{[P37]} \textit{When Users Don't Ask: Benchmarking Context-Driven Memory Retrieval in Conversational Agents}~\citep{chang-chen-2026-users-dont-ask}; \textbf{[P44]} \textit{UTILMEM: Benchmarking Evidence Utilization in Long-Term Conversational Memory}~\citep{qing-etal-2026-utilmem}; \textbf{[P5]} \textit{PRAGMA: Evaluating Personalized Guidance with Memory Alignment in Lifelong Conversations}~\citep{yu-etal-2026-pragma}\\
\end{longtable}
\endgroup

\begingroup\small
\setlength{\tabcolsep}{6pt}
\renewcommand{\arraystretch}{1.15}
\begin{longtable}{>{\raggedright\arraybackslash}p{\dimexpr\textwidth-2\tabcolsep\relax}}
\toprule
\textbf{Generated output:} Expert B insight --- Evolving Memory Architecture\\
\midrule\endfirsthead
\toprule
\textbf{Generated output:} Expert B insight --- Evolving Memory Architecture (continued)\\
\midrule\endhead
\midrule\hfill Continued on next page\\\endfoot
\bottomrule\noalign{\vskip 6pt}
\caption{Insight distilled for Expert~B (Evolving Memory Architecture), reproduced verbatim. Shaded rows are model output; the paper key below them is added by the authors.}\label{tab:case-insight-b}\\
\endlastfoot
\rowcolor{gray!12}\textbf{Core insight.} This line keeps running into the fact that static summaries and independent top-k retrieval cannot represent changing truths, provenance, conflicts, and query-specific evidence needs together. Systems have progressed from profile-memory graphs to temporal state graphs, multi-round retrieval, selective memory types, and lifecycle-aware relations. They assume explicit structure can make memory auditable without retaining the full history. The unresolved cost is deciding which structure to maintain and when updates or retrieval should reshape it.\\
\rowcolor{gray!12}\textbf{Historical tension.} More explicit temporal and evidential structure improves correctness and traceability, but increases maintenance, retrieval complexity, and dependence on intermediate decisions.\\
\rowcolor{gray!12}\textbf{Grounding summary.} This reading rests on P10, P9, P24, P11, and P30, the graph's densest line of structured, evolving, and lifecycle-aware memory systems. It shows a progression from organizing memory to explicitly governing evidence state and retrieval-driven maintenance.\\
\rowcolor{gray!12}\textbf{Supporting findings.}\\
\rowcolor{gray!12}\hspace*{1em}\textbf{[P10]} PGMem linked persona nodes to event nodes through typed provenance and evidence edges, consistently outperforming summary-based, persona-aware, graph, and agentic baselines across three benchmarks.\\
\rowcolor{gray!12}\hspace*{1em}\textbf{[P9]} EMIR\textsuperscript{2} used a State-Evolving Memory Graph and intent-guided multi-round retrieval, producing relative improvements exceeding 12\% in some LoCoMo and MemConflict categories.\\
\rowcolor{gray!12}\hspace*{1em}\textbf{[P24]} LeanMem routed information into profile, event, or source-grounded record memory; on LoCoMo and LongMemEval-S it improved over the strongest memory baseline by up to 15.1 points.\\
\rowcolor{gray!12}\hspace*{1em}\textbf{[P11]} RIME used self-questions to retrieve evidence during memory formation and achieved the best results across all three reported quality metrics on LoCoMo with Qwen3-235B-A22B and GPT-5.6 Sol.\\
\rowcolor{gray!12}\hspace*{1em}\textbf{[P30]} MemoryLACE modeled merge, supersession, and contradiction relations while preserving atomic memories; it reduced BEAM end-to-end runtime by 66.6\% versus Hindsight.\\
\midrule
\footnotesize\textit{Paper key.} \textbf{[P10]} \textit{PGMem: Tightly Coupled Persona-Memory Graph for Lifelong Personalized Agents}~\citep{choi-etal-2026-pgmem}; \textbf{[P9]} \textit{EMIR\textsuperscript{2}: Evolution-Aware Memory with Intent-Guided Multi-Round Retrieval}~\citep{liu-etal-2026-emir2}; \textbf{[P24]} \textit{LeanMem: Simple and Efficient Long-Term Memory for LLM Agents}~\citep{liao-etal-2026-leanmem}; \textbf{[P11]} \textit{Remember by Asking: Retrieval-Induced Memory Evolution for LLM Agents}~\citep{zhou-etal-2026-rime}; \textbf{[P30]} \textit{MemoryLACE: Memory Lifecycle-Aware Consolidation and Evidence Retrieval}~\citep{yacoubi-etal-2026-memorylace}\\
\end{longtable}
\endgroup

\begingroup\small
\setlength{\tabcolsep}{6pt}
\renewcommand{\arraystretch}{1.15}
\begin{longtable}{>{\raggedright\arraybackslash}p{\dimexpr\textwidth-2\tabcolsep\relax}}
\toprule
\textbf{Generated output:} Expert C insight --- Continual Personalization Alignment\\
\midrule\endfirsthead
\toprule
\textbf{Generated output:} Expert C insight --- Continual Personalization Alignment (continued)\\
\midrule\endhead
\midrule\hfill Continued on next page\\\endfoot
\bottomrule\noalign{\vskip 6pt}
\caption{Insight distilled for Expert~C (Continual Personalization Alignment), reproduced verbatim. Shaded rows are model output; the paper key below them is added by the authors.}\label{tab:case-insight-c}\\
\endlastfoot
\rowcolor{gray!12}\textbf{Core insight.} This line keeps finding that personalization signals are latent, noisy, time-varying, and easy to over-apply. It has moved from static profiles toward uncertainty-aware belief revision, preference hypotheses, adaptive stopping, and online tracing of short- and long-term intent. It assumes alignment should reflect inferred user constraints rather than merely repeat stored preferences. It remains stuck between adapting quickly enough to matter and resisting transient evidence, preference narrowing, sycophancy, and irrelevant personalization.\\
\rowcolor{gray!12}\textbf{Historical tension.} Personalization must be responsive to emerging preferences without turning every weak signal into a persistent commitment or an excuse to flatter the user.\\
\rowcolor{gray!12}\textbf{Grounding summary.} This reading rests on P19, P16, P17, P42, P47, and P46, spanning personalization-risk evaluation, behavioral alignment, belief revision, hypothesis tracking, and context-budget control. The graph shows adaptation becoming more explicit and selective, but not yet reliably safe.\\
\rowcolor{gray!12}\textbf{Supporting findings.}\\
\rowcolor{gray!12}\hspace*{1em}\textbf{[P19]} Across 13 LLMs, PRISK found personalization increased irrelevant personalization, preference narrowing, and sycophantic bias, with reported average drops of 45.9\%, 41.7\%, and 61.7\%, respectively.\\
\rowcolor{gray!12}\hspace*{1em}\textbf{[P16]} IBA-Bench showed a knowledge-to-action gap: agents must execute tasks under implicit, noisy, temporally inconsistent constraints, and IBA-Agent improved behavioral alignment across nine application domains.\\
\rowcolor{gray!12}\hspace*{1em}\textbf{[P17]} CORE separated turn-local evidence from persistent persona revision through uncertainty-aware belief updates, improving alignment and robustness on ALOE, PersonaChat, and PERSIST.\\
\rowcolor{gray!12}\hspace*{1em}\textbf{[P42]} HypReflect represented uncertain preference hypotheses, refined them as evidence accumulated, and used hypothesis-guided self-distillation across online, multi-session, and implicit-behavior settings.\\
\rowcolor{gray!12}\hspace*{1em}\textbf{[P47]} HyperTrace maintained interpretable short-term-intent and long-term-preference hypotheses with an SMC-style reweighting process, improving alignment, preference prediction, and profile consistency on PRISM and PersonaMem-v2.\\
\rowcolor{gray!12}\hspace*{1em}\textbf{[P46]} ENOUGH learned to append records or stop, optimizing downstream utility, user specificity, and token cost to construct minimal sufficient profiles across six personalized tasks.\\
\midrule
\footnotesize\textit{Paper key.} \textbf{[P19]} \textit{Evaluating the Hidden Costs of Personalization in Large Language Models}~\citep{wang-etal-2026-hidden-costs}; \textbf{[P16]} \textit{From Profiling to Synthesis: Benchmarking Implicit Behavioral Alignment in Personalized LLM Agents}~\citep{song-etal-2026-iba-bench}; \textbf{[P17]} \textit{Toward Robust Personalized Alignment for LLMs: Mitigating Persona Drift in Multi-Turn Dialogue}~\citep{zhang-etal-2026-persona-drift}; \textbf{[P42]} \textit{Hypotheses-Guided Self Distillation for Continual Personalization}~\citep{hwang-etal-2026-hypreflect}; \textbf{[P47]} \textit{HyperTrace: Hypothesis-Based Preference Tracing for Online LLM Personalization}~\citep{shen-etal-2026-hypertrace}; \textbf{[P46]} \textit{Less Is Personal: Learning Minimal Sufficient User Profiles for Personalized Language Models}~\citep{liu-etal-2026-less-is-personal}\\
\end{longtable}
\endgroup

%% file: tables/case_proposal.tex
\begingroup\small
\setlength{\tabcolsep}{6pt}
\renewcommand{\arraystretch}{1.15}
\begin{longtable}{>{\raggedright\arraybackslash}p{\dimexpr\textwidth-2\tabcolsep\relax}}
\toprule
\textbf{Generated output:} Final research proposal\\
\midrule\endfirsthead
\toprule
\textbf{Generated output:} Final research proposal (continued)\\
\midrule\endhead
\midrule\hfill Continued on next page\\\endfoot
\bottomrule\noalign{\vskip 6pt}
\caption{Final proposal produced by Stage~3, reproduced verbatim (all reader-facing fields; the internal expert-attribution field is omitted). Long fields are split into rows at sentence boundaries for page layout only.}\label{tab:case-proposal}\\
\endlastfoot
\rowcolor{gray!12}\textbf{Title.} Lifecycle-Aware Counterfactual Control of Memory Influence\\
\rowcolor{gray!12}\textbf{Problem and motivation.} Can a personalized dialogue system determine which evolving user-memory evidence is currently justified, whether it applies to the present request, and how strongly it should change the response? A memory may be current, historical, context-specific, superseded, unresolved, or irrelevant. The appropriate behavior may therefore be silent personalization, explicit grounding, confirmation, abstention, or no behavioral change. Existing evaluations often reduce personalization to retrieval accuracy, answer correctness, or explicit memory mention.\\
\rowcolor{gray!12}This misses useful silent grounding and cannot distinguish missing evidence from a system that retrieved the right evidence but gave it unjustified authority. The proposal studies whether lifecycle-aware evidence states and auditable influence control can produce more useful and less harmful personalization in recommendation, planning, and decision-support tasks.\\
\rowcolor{gray!12}\textbf{Method.} The system constructs a lifecycle-aware evidence unit from retrieved memories. It represents whether claims are current, historical, context-specific, superseded, unresolved, or irrelevant, together with provenance and supporting or conflicting relations. A lightweight controller receives the current request, this evidence unit, and a no-memory shadow response. It selects an action---silent use, explicit grounding, confirmation, or abstention---and one of four discrete influence levels. Level 0 prevents memory from changing ranking, membership, or plan constraints. Level 0.25 permits reranking only among options that already satisfy the explicit task constraints.\\
\rowcolor{gray!12}Level 0.5 permits a soft preference or constraint while retaining at least one viable alternative. Level 1 permits filtering options or materially changing a plan only when the evidence is a confirmed current constraint, such as an explicit recent statement or concordant evidence meeting a preregistered rule. If level 1 is predicted without confirmation, the system downgrades to level 0.5 or asks for confirmation. The controller applies the intervention to a common candidate pool or structured plan representation; the generator then receives only that representation, not the retrieved evidence or controller action, and verbalizes it with deterministic decoding.\\
\rowcolor{gray!12}Confirmation produces only a clarification question, explicit grounding must accurately mention the relevant evidence, and abstention uses the no-memory response. Influence is measured primarily through changes in option ranking, membership, retained alternatives, and plan constraints, and secondarily through textual grounding, usefulness, and naturalness. This makes memory influence an auditable intervention rather than a latent score or a proxy based on memory mention.\\
\rowcolor{gray!12}\textbf{Experiment.} The primary study will evaluate structured recommendation and planning cases drawn from eligible examples in MemUse, When Users Don't Ask: Benchmarking Context-Driven Memory Retrieval in Conversational Agents, UTILMEM: Benchmarking Evidence Utilization in Long-Term Conversational Memory, and PRAGMA: Evaluating Personalized Guidance with Memory Alignment in Lifelong Conversations. A preliminary audit will determine whether these resources contain sufficient suitable instances; if not, the natural-data component will be reported as exploratory rather than presented as representative. The benchmark will have two separated components.\\
\rowcolor{gray!12}The controlled component will contain approximately 50 cases for each of five conditions: supersession, contextual coexistence, contradiction, stale distractors, and irrelevance. These labels are known by construction, but cases with the same lifecycle state will vary in request risk and applicability so that state classification alone cannot solve the task. The natural component will use unmodified or minimally transformed histories where possible: 300 cases for development and calibration and 100 held-out cases for final evaluation. Three annotators will label each case's lifecycle state, acceptable action set, maximum permitted influence, and whether the correct behavior is no change.\\
\rowcolor{gray!12}They will not label system outputs. Three blinded raters will independently judge held-out outputs against the same case and its no-memory reference for useful improvement, harmful degradation, grounding accuracy, and naturalness. Agreement will be reported before adjudication, and mixed-effects or paired-bootstrap intervals will account for case and rater clustering. All systems will use the same request, history, retrieved evidence, candidate pool or initial plan, prompt template, model, decoding settings, and random seed. The no-memory and memory branches will differ only in the prescribed intervention, and verbalization will receive only the resulting structured representation.\\
\rowcolor{gray!12}Baselines will include no memory, aggressive top-k retrieval, lifecycle retrieval without influence control, influence control without lifecycle relations, a confidence-only controller, and random action or budget policies calibrated on development data to match abstention and mean influence. The primary estimand is the held-out paired difference in useful influence and harmful influence between the complete method and the strongest matched control. The preregistered success criterion is at least a 0.10-point improvement in useful influence without more than a 0.05-point increase in harmful influence on the rating scale.\\
\rowcolor{gray!12}Additional measures are lifecycle-state accuracy, action-set accuracy, influence-level accuracy, no-change accuracy, missed and unnecessary confirmation, and explicit-grounding correctness. A result supporting the claim must show higher useful and lower harmful behavioral change at matched abstention and average influence, with lifecycle ablations failing especially on supersession, contextual coexistence, and contradiction cases. A null result will be interpreted narrowly as non-improvement for this controller, benchmark, and task domain.\\
\rowcolor{gray!12}If the primary result holds, a follow-up study can test longitudinal authority decay and recommendation-induced self-confirmation; these are not claims of the present paper.\\
\rowcolor{gray!12}\textbf{Novelty.} The proposal makes memory's permitted behavioral influence---not retrieval, explicit mention, or binary utilization---the central object of study. It combines lifecycle-aware evidence reconstruction with a discrete, auditable policy over how much that evidence may change a recommendation or plan. The no-memory shadow response exposes silent adaptation, while fixed intervention levels make behavioral changes attributable and testable. The design also treats justified no-change behavior as a first-class outcome and controls for improvements caused merely by retrieving more evidence, abstaining more often, or changing prompt format.\\
\rowcolor{gray!12}\textbf{Builds on.} MemUse: Moving Memory Evaluation from Direct QA to Natural Integration in Long-Term Human-AI Conversation (2026) reports that a system can perform well on direct fact questions while naturally referencing very few corresponding facts, and that natural integration better tracks satisfaction; this motivates evaluating behavioral change and silent grounding rather than explicit memory mention. MemoryLACE: Memory Lifecycle-Aware Consolidation and Evidence Retrieval (2026) motivates organizing evidence through lifecycle relations such as merge, supersession, and contradiction before using it; this proposal uses that lifecycle state as the input to its influence controller.\\
\rowcolor{gray!12}PRAGMA: Evaluating Personalized Guidance with Memory Alignment in Lifelong Conversations (2026) motivates testing memory use in evolving personalized guidance, where incorrect assumptions can alter recommendations and plans rather than merely affect factual recall.\\
\rowcolor{gray!12}\textbf{Distinct from.} Unlike MemUse: Moving Memory Evaluation from Direct QA to Natural Integration in Long-Term Human-AI Conversation (2026), which establishes the importance of natural integration but does not provide matched interventions for attributing whether memory changed behavior appropriately, this proposal compares paired no-memory, justified-memory, stale-memory, and conflict conditions and evaluates the resulting structured and textual changes.\\
\rowcolor{gray!12}Unlike MemoryLACE: Memory Lifecycle-Aware Consolidation and Evidence Retrieval (2026), which organizes and retrieves memories using lifecycle relations, this proposal studies the subsequent governance decision: whether that evidence should silently change behavior, be stated, trigger confirmation, or be withheld, and how much influence it may exert.\\
\rowcolor{gray!12}\textbf{Potential impact.} If successful, the work will give personalized-dialogue research a measurable distinction between possessing a justified memory and granting it justified behavioral authority. It will help separate missing evidence, incorrect lifecycle interpretation, excessive influence, ignored evidence, and generation failures, while supporting useful personalization without treating every retrieved fact as an instruction. Its fixed interventions and matched controls provide a practical evaluation and control layer for public-model systems under university-lab compute.\\
\rowcolor{gray!12}If it does not improve human-rated usefulness and harm at matched abstention and influence levels, the result will show that this lifecycle-aware controller does not improve personalization under the tested conditions, or that generation and user modeling remain the dominant bottlenecks; it will not by itself establish that memory authority is generally invalid.\\
\end{longtable}
\endgroup

%% file: tables/case_initial.tex
\begingroup\small
\setlength{\tabcolsep}{6pt}
\renewcommand{\arraystretch}{1.15}
\begin{longtable}{>{\raggedright\arraybackslash}p{0.18\textwidth}>{\columncolor{gray!12}\raggedright\arraybackslash}p{\dimexpr0.82\textwidth-4\tabcolsep\relax}}
\toprule
\multicolumn{2}{l}{\textbf{Generated output:} Expert proposals in the first debate turn}\\
\midrule\endfirsthead
\toprule
\multicolumn{2}{l}{\textbf{Generated output:} Expert proposals in the first debate turn (continued)}\\
\midrule\endhead
\midrule\multicolumn{2}{l}{\hfill Continued on next page}\\\endfoot
\bottomrule\noalign{\vskip 6pt}
\caption{Excerpts from the three expert proposals in Stage~2, turn~1 (speaking order C, A, B), reproduced verbatim.}\label{tab:case-initial}\\
\endlastfoot
\textit{Expert A, turn 1} & ``A system receives credit when the memory-induced change is useful and warranted---even if the response never mentions the memory.''\\
\textit{Expert C, turn 1} & ``Rather than passing retrieved memories directly to a generator, IGP produces an \textbf{influence budget} for each hypothesis.''\\
\textit{Expert B, turn 1} & ``From the evolving-memory perspective, however, influence cannot be calibrated reliably if the input to the controller is merely a bag of retrieved memories with confidence scores. Confidence is not a substitute for lifecycle state.''\\
\end{longtable}
\endgroup

%% file: tables/case_chain_budget.tex
\begingroup\small
\setlength{\tabcolsep}{6pt}
\renewcommand{\arraystretch}{1.15}
\begin{longtable}{>{\raggedright\arraybackslash}p{0.18\textwidth}>{\columncolor{gray!12}\raggedright\arraybackslash}p{\dimexpr0.82\textwidth-4\tabcolsep\relax}}
\toprule
\multicolumn{2}{l}{\textbf{Generated output:} Revision chain 1 --- making the influence budget operational}\\
\midrule\endfirsthead
\toprule
\multicolumn{2}{l}{\textbf{Generated output:} Revision chain 1 --- making the influence budget operational (continued)}\\
\midrule\endhead
\midrule\multicolumn{2}{l}{\hfill Continued on next page}\\\endfoot
\bottomrule\noalign{\vskip 6pt}
\caption{Excerpts tracing the influence budget from the first draft to the final proposal, reproduced verbatim.}\label{tab:case-chain-budget}\\
\endlastfoot
\textit{Draft, turn 1} & ``The budget determines how strongly a memory may alter the response: a confirmed current constraint may substantially change a plan, while a weak inferred preference may only softly rerank options.''\\
\textit{Critic, after turn 1} & ``Otherwise the claimed budget may merely be a latent label and the observed behavior could be caused by generation noise or prompt wording.''\\
\textit{Expert C, turn 2} & ``The system should first generate a common candidate pool or structured plan representation. The influence gate then modifies that representation according to the budget, and the same frozen generator verbalizes the result.''\\
\textit{Draft, turn 2} & ``Level 0 excludes memory from candidate selection and planning; level 0.25 permits reranking compatible options without removing alternatives; level 0.5 permits a soft constraint while retaining alternatives; and level 1 permits filtering or materially changing a plan only when the evidence is a confirmed current constraint.''\\
\textit{Final proposal} & ``The controller applies the intervention to a common candidate pool or structured plan representation; the generator then receives only that representation, not the retrieved evidence or controller action, and verbalizes it with deterministic decoding.''\\
\end{longtable}
\endgroup

%% file: tables/case_chain_decay.tex
\begingroup\small
\setlength{\tabcolsep}{6pt}
\renewcommand{\arraystretch}{1.15}
\begin{longtable}{>{\raggedright\arraybackslash}p{0.18\textwidth}>{\columncolor{gray!12}\raggedright\arraybackslash}p{\dimexpr0.82\textwidth-4\tabcolsep\relax}}
\toprule
\multicolumn{2}{l}{\textbf{Generated output:} Revision chain 2 --- aligning the claim with the evaluation scope}\\
\midrule\endfirsthead
\toprule
\multicolumn{2}{l}{\textbf{Generated output:} Revision chain 2 --- aligning the claim with the evaluation scope (continued)}\\
\midrule\endhead
\midrule\multicolumn{2}{l}{\hfill Continued on next page}\\\endfoot
\bottomrule\noalign{\vskip 6pt}
\caption{Excerpts tracing the deferred authority-decay mechanism, reproduced verbatim.}\label{tab:case-chain-decay}\\
\endlastfoot
\textit{Draft, turn 1} & ``Repeated use of an uncertain hypothesis reduces its authority unless new independent evidence supports it, preventing the system's own recommendations from becoming false confirmation.''\\
\textit{Critic, after turn 1} & ``The proposed primary benchmark appears largely case-based, while this rule requires multi-turn trajectories and outcomes; either add a fully specified longitudinal protocol and baselines or remove the rule from the primary claim.''\\
\textit{Expert C, turn 2} & ``The longitudinal authority-decay mechanism is interesting, but it currently asks for a different paper: multi-turn exposure, user feedback, and a model of independence between evidence sources.''\\
\textit{Final proposal} & ``If the primary result holds, a follow-up study can test longitudinal authority decay and recommendation-induced self-confirmation; these are not claims of the present paper.''\\
\end{longtable}
\endgroup

%% file: tables/prompt_directory.tex
\begingroup\small
\setlength{\tabcolsep}{5pt}
\renewcommand{\arraystretch}{1.15}
\begin{longtable}{@{}p{0.24\textwidth}p{0.36\textwidth}p{\dimexpr0.40\textwidth-20pt\relax}@{}}
\toprule
\textbf{Workflow group} & \textbf{Prompts} & \textbf{Input $\rightarrow$ output} \\
\midrule
\hyperref[sec:prompts-expertise]{\S\ref*{sec:prompts-expertise} Expertise construction}
& Seed selection; related-paper filtering; insight distillation
& Research topic and literature $\rightarrow$ paper graph and three evidence-supported insights \\
\addlinespace
\hyperref[sec:prompts-setup]{\S\ref*{sec:prompts-setup} Agent setup}
& Expert and moderator system messages
& Insight and supporting papers $\rightarrow$ expert context; standing rules for moderation \\
\addlinespace
\hyperref[sec:prompts-problem]{\S\ref*{sec:prompts-problem} Problem selection}
& Problem statements; anonymous shortlist; blind ranking; final selection
& Expert perspectives $\rightarrow$ one shared research problem \\
\addlinespace
\hyperref[sec:prompts-debate]{\S\ref*{sec:prompts-debate} Proposal development}
& Expert proposals; first draft; critic review; revision; final positions; draft update
& Selected problem and expert knowledge $\rightarrow$ a revised proposal \\
\addlinespace
\hyperref[sec:prompts-synthesis]{\S\ref*{sec:prompts-synthesis} Final synthesis}
& Final proposal write-up
& Discussion and latest draft $\rightarrow$ one structured research idea \\
\addlinespace
\hyperref[sec:prompts-evaluation]{\S\ref*{sec:prompts-evaluation} Idea evaluation}
& Pairwise judge, including the five evaluation criteria
& Topic and two proposals $\rightarrow$ a preference and rationale for each criterion \\
\bottomrule
\noalign{\vskip 6pt}
\caption{Prompt directory. The first five groups generate a research idea; the last evaluates completed ideas. Group names link to the corresponding appendix subsections.}
\label{tab:prompt-directory}\\
\end{longtable}
\endgroup

%% file: tables/prompts_graph.tex
\begingroup\small
\setlength{\tabcolsep}{0pt}
\renewcommand{\arraystretch}{1.08}
\begin{longtable}{@{}>{\raggedright\arraybackslash}p{\textwidth}@{}}
\toprule
\textbf{Seed-paper selection}\\
\midrule\endfirsthead
\toprule
\textbf{Seed-paper selection} (continued)\\
\midrule\endhead
\midrule\hfill Continued on next page\\\endfoot
\bottomrule\noalign{\vskip 6pt}
\caption{Seed-paper selection}\label{tab:prompt-seed}\\
\endlastfoot
\textit{Reading note.} Used once to select a starting paper. Candidate papers contain numbered titles and abstracts.\\[5pt]
\textbf{System message}\\*[2pt]
You pick the starting paper for a research graph about a topic.\\[4pt]
You get a topic and a numbered list of recent papers that a search returned for it. Search ranks by word overlap, so the top hit is often a paper from a different research area that happens to share the topic's words. Your job is to pick the one paper that is really about the topic.\\[4pt]
Rules:\newline 1. The paper must be about the topic itself, not merely share words with it. If a term in the topic means something different in that paper's field, it is the wrong pick.\newline 2. Prefer a paper whose main problem IS the topic over one that only touches it as a component.\newline 3. Prefer a paper that sits in the middle of the topic's literature over a narrow or unusual application of it --- the whole graph is grown from this paper.\newline 4. Citations, venue and date only break ties.\newline 5. Pick exactly one, even if none is a perfect fit; say so in `why` when the fit is poor.\\[4pt]
Return ONLY this JSON object, no other text:\newline \{"pick": \textless{}candidate number\textgreater{}, "why": "\textless{}at most 20 words\textgreater{}"\}\\[4pt]
\addlinespace[4pt]
\textbf{User message}\\*[2pt]
TOPIC: [Research topic]\\[4pt]
CANDIDATES:\newline [Candidate papers]\\[4pt]
Pick the one paper this topic's research graph should be grown from. Return the JSON object described above.\\[4pt]
\addlinespace[4pt]
\end{longtable}
\endgroup

\begingroup\small
\setlength{\tabcolsep}{0pt}
\renewcommand{\arraystretch}{1.08}
\begin{longtable}{@{}>{\raggedright\arraybackslash}p{\textwidth}@{}}
\toprule
\textbf{Related-paper filtering}\\
\midrule\endfirsthead
\toprule
\textbf{Related-paper filtering} (continued)\\
\midrule\endhead
\midrule\hfill Continued on next page\\\endfoot
\bottomrule\noalign{\vskip 6pt}
\caption{Related-paper filtering}\label{tab:prompt-neighbors}\\
\endlastfoot
\textit{Reading note.} The first expansion level is shown with a limit of 5 papers per parent. The same system message uses 3 and 2 at the second and third levels. Source paper and candidate papers include metadata and abstracts.\\[5pt]
\textbf{System message}\\*[2pt]
You filter paper recommendations for a research topic.\\[4pt]
You get: a topic, one source paper, and a numbered list of papers that Semantic Scholar recommends for that source paper. Recommendations are noisy --- many are only loosely related. Keep the ones worth adding to a research graph about the topic.\\[4pt]
Rules:\newline 1. Relevance to the topic comes first. A paper that is close to the source paper but off-topic is not worth keeping.\newline 2. The paper must also be a real neighbour of the source paper's line of work: it addresses the same problem, uses or replaces the same kind of method, or evaluates the same thing.\newline 3. Prefer papers that add something the graph does not have yet --- a different method, setting, or failure mode --- over near-duplicates of the source paper or of each other.\newline 4. Keep at most 5. Keeping fewer is correct when fewer are relevant; keeping none is allowed.\\[4pt]
Return ONLY this JSON object, no other text:\newline \{"selected": [\{"id": \textless{}candidate number\textgreater{}, "why": "\textless{}at most 15 words\textgreater{}"\}]\}\\[4pt]
\addlinespace[4pt]
\textbf{User message}\\*[2pt]
TOPIC: [Research topic]\\[4pt]
SOURCE PAPER:\newline [Source paper]\\[4pt]
CANDIDATES:\newline [Candidate papers]\\[4pt]
Select at most 5 candidates. Return the JSON object described above.\\[4pt]
\addlinespace[4pt]
\end{longtable}
\endgroup

\begingroup\small
\setlength{\tabcolsep}{0pt}
\renewcommand{\arraystretch}{1.08}
\begin{longtable}{@{}>{\raggedright\arraybackslash}p{\textwidth}@{}}
\toprule
\textbf{Insight distillation}\\
\midrule\endfirsthead
\toprule
\textbf{Insight distillation} (continued)\\
\midrule\endhead
\midrule\hfill Continued on next page\\\endfoot
\bottomrule\noalign{\vskip 6pt}
\caption{Insight distillation}\label{tab:prompts-graph}\\
\endlastfoot
\textit{Reading note.} One call reads the whole graph to produce 3 insights, each supported by at least 2 papers. Graph papers contain identifiers, metadata and abstracts; graph relations use those identifiers. The numbers and provenance paragraph are filled for the online configuration, where an edge links two papers that Semantic Scholar recommends as similar and its direction follows publication date only. The transition summary is the model's interpretation and is not shown to the experts.\\[5pt]
\textbf{System message}\\*[2pt]
You read a graph of related papers and write 3 pieces of background knowledge from it --- the understanding a researcher has after living inside one line of work for a while.\\[4pt]
The graph was grown from one recent paper: each round asked Semantic Scholar for related papers and kept the ones about the topic. Most of it sits near the seed, but the graph usually also holds papers from other kinds of work on the topic --- training it in, steering it inside the model, watching it in agents, building data for it. Those matter as much as the dense centre.\\[4pt]
FIRST, SORT THE GRAPH\\[4pt]
Before writing anything, sort the papers into the distinct lines of work the graph contains. A line of work is a kind of research --- measuring something, training for it, locating it inside models, studying it in deployed agents, building the data --- not a different angle on one question. Then choose the 3 most different lines that have at least two papers each. If the graph only carries fewer than 3 kinds of work, the remaining insight may be a second line inside the largest kind; say so in grounding\_\allowbreak{}summary.\\[4pt]
WHAT YOU ARE WRITING\\[4pt]
Each insight is one line of work seen from the inside: what problem it keeps running into, what has been tried against it and how that has shifted, what it takes for granted, and where it is stuck. Its purpose is to inform someone who will then think freely about the topic --- it should open up what they can see, not hand them a move to reuse.\\[4pt]
Alongside the reading, keep the payload: the specific findings the stance rests on, with the method names, benchmarks and numbers as reported. The reading is what an insider believes; the findings are why they believe it --- and they are the one part of expertise a newcomer cannot invent.\\[4pt]
Two versions of an insight about the same three papers on retrieval-augmented generation:\newline \hspace*{0.8em}Recipe: "Retrievers are repeatedly fine-tuned on downstream task signal to close the gap between retrieval relevance and answer usefulness."\newline \hspace*{0.8em}Background: "This line of work keeps finding that what a retriever ranks highly is not what the generator needs --- the two are trained on different notions of relevance. Closing the gap by training the retriever on the generator's signal has helped but couples the two tightly, and nobody has yet asked whether the generator should instead learn to use imperfect retrieval."\newline The second tells a reader where the frontier is and leaves them free to decide what to do about it.\\[4pt]
Write each insight for a researcher from a different sub-field: name the line the way the field would name it, say what it keeps finding and where it is stuck in plain words, and keep the method names, benchmarks and numbers the papers report --- saying in a few words what each named thing is, so a newcomer can use the finding without having read the paper. The papers are your evidence for what the line has found; the insight is about the line, not about the papers.\\[4pt]
OUTPUT\\[4pt]
Return only this JSON, using the [P\#] labels exactly as given:\\[4pt]
\{\newline \hspace*{0.8em}"insights": [\newline \hspace*{1.6em}\{\newline \hspace*{2.4000000000000004em}"insight\_\allowbreak{}name": "2-5 word name for this line of work, as the field would say it",\newline \hspace*{2.4000000000000004em}"core\_\allowbreak{}insight": "3-4 sentences: what this line keeps running into, what has been tried, what it assumes, where it is stuck (max 90 words)",\newline \hspace*{2.4000000000000004em}"historical\_\allowbreak{}tension": "the trade-off or pressure that keeps it moving (max 40 words)",\newline \hspace*{2.4000000000000004em}"claims": [\\[4pt]
\hspace*{3.2em}\{"label": "P3",\newline \hspace*{4.0em}"finding": "one specific result that paper reports and this reading rests on, with the names and numbers it uses (max 35 words)"\}\newline \hspace*{2.4000000000000004em}],\newline \hspace*{2.4000000000000004em}"grounding\_\allowbreak{}summary": "which papers this reading rests on, and how much of this line the graph actually shows (max 60 words)",\newline \hspace*{2.4000000000000004em}"papers": [\newline \hspace*{3.2em}\{"label": "P3",\newline \hspace*{4.0em}"neutral\_\allowbreak{}description": "what this paper did, factually (max 25 words)",\\[4pt]
\hspace*{4.0em}"role\_\allowbreak{}in\_\allowbreak{}insight": "its place in this line of work (max 25 words)"\}\newline \hspace*{2.4000000000000004em}],\newline \hspace*{2.4000000000000004em}"edges": [\newline \hspace*{3.2em}\{"edge\_\allowbreak{}id": "P3\(\rightarrow\)P7",\newline \hspace*{4.0em}"transition\_\allowbreak{}summary": "what changed between the two papers and why (max 30 words)",\newline \hspace*{4.0em}"role\_\allowbreak{}in\_\allowbreak{}abstraction": "what this step shows about the line of work (max 25 words)"\}\newline \hspace*{2.4000000000000004em}]\\[4pt]
\hspace*{1.6em}\}\newline \hspace*{0.8em}]\newline \}\\[4pt]
Every insight rests on at least 2 papers that are actually in the graph; list the edges between them that exist in the EDGES list, and leave "edges" empty when there are none --- a line of work at the graph's edge often has papers but no recorded steps. Use only [P\#] labels and edge ids from the lists. Give each insight 4-6 claims, each from a different paper where the evidence allows it. Don't invent papers, methods or results.\\[4pt]
\addlinespace[4pt]
\textbf{User message}\\*[2pt]
TOPIC: [Research topic]\\[4pt]
PAPERS IN THE GRAPH:\newline [Graph papers]\\[4pt]
EDGES (older paper \(\rightarrow\) newer paper):\newline [Graph relations]\\[4pt]
Write 3 different insights from this graph. Return the JSON object described above.\\[4pt]
\addlinespace[4pt]
\end{longtable}
\endgroup

%% file: tables/prompts_roles.tex
\begingroup\small
\setlength{\tabcolsep}{0pt}
\renewcommand{\arraystretch}{1.08}
\begin{longtable}{@{}>{\raggedright\arraybackslash}p{\textwidth}@{}}
\toprule
\textbf{Expert setup}\\
\midrule\endfirsthead
\toprule
\textbf{Expert setup} (continued)\\
\midrule\endhead
\midrule\hfill Continued on next page\\\endfoot
\bottomrule\noalign{\vskip 6pt}
\caption{Expert setup}\label{tab:prompts-roles}\\
\endlastfoot
\textit{Reading note.} Persistent system message for problem statements and proposal discussion. The blind vote uses a separate neutral voting prompt.\\[5pt]
\textbf{System message}\\*[2pt]
You are one of three researchers working out a research proposal together. The goal is one idea good enough for a top ML venue: new, and worth doing.\\[4pt]
\textbf{Where you are coming from}\\*[4pt]
You have spent the last while inside one line of work, and it left you with a way of seeing the field:\\[4pt]
\textit{[Expertise block: this expert's insight name, research perspective and historical tension; supporting findings linked to source papers; and evidence-paper identifiers, titles, authors, years, venues when available, and abstracts. The block also notes that only abstracts, not full papers, are available.]}\\[4pt]
This is where you are coming from, and it is what makes your take different from the other two. Look at the problem from inside this line of work first --- what does it see that the others miss? Then bring in whatever else you know: your own knowledge of the field, methods from elsewhere, ideas that have nothing to do with these papers.\\[4pt]
\textbf{How to be useful here}\\*[4pt]
Aim at the scale a paper works at: something the field would learn from, build on or measure with --- not an adjustment to one component. Propose, sharpen, combine. When another researcher's idea is stronger than yours, build on it rather than defending your own.\\[4pt]
Say when something is established and when it is your own extrapolation. Don't invent papers or results.\\[4pt]
\addlinespace[4pt]
\end{longtable}
\endgroup

\begingroup\small
\setlength{\tabcolsep}{0pt}
\renewcommand{\arraystretch}{1.08}
\begin{longtable}{@{}>{\raggedright\arraybackslash}p{\textwidth}@{}}
\toprule
\textbf{Moderator setup}\\
\midrule\endfirsthead
\toprule
\textbf{Moderator setup} (continued)\\
\midrule\endhead
\midrule\hfill Continued on next page\\\endfoot
\bottomrule\noalign{\vskip 6pt}
\caption{Moderator setup}\label{tab:prompt-moderator}\\
\endlastfoot
\textit{Reading note.} Persistent system message for shortlisting, selection, drafting, revision and final synthesis. The moderator receives the substantive discussion records available at each step.\\[5pt]
\textbf{System message}\\*[2pt]
You are running a discussion among three researchers and writing up what it produces: one research proposal.\\[4pt]
Stage 1 settles which problem to work on; once settled it stays settled. Stage 2 develops a draft through several rounds. Stage 3 is the final write-up.\\[4pt]
Your write-ups are built from what the researchers said --- you choose, combine and sharpen, you don't add methods or claims of your own. Within that, be ambitious on their behalf: the proposal should be the strongest paper their ideas support, not the safest.\\[4pt]
\addlinespace[4pt]
\end{longtable}
\endgroup

%% file: tables/prompts_problem.tex
\begingroup\small
\setlength{\tabcolsep}{0pt}
\renewcommand{\arraystretch}{1.08}
\begin{longtable}{@{}>{\raggedright\arraybackslash}p{\textwidth}@{}}
\toprule
\textbf{Stage 1: problem statements}\\
\midrule\endfirsthead
\toprule
\textbf{Stage 1: problem statements} (continued)\\
\midrule\endhead
\midrule\hfill Continued on next page\\\endfoot
\bottomrule\noalign{\vskip 6pt}
\caption{Stage 1: problem statements}\label{tab:prompt-problem}\\
\endlastfoot
\textit{Reading note.} Each expert states an open problem using its standing system message. The instruction below is appended to the discussion history available to that expert.\\[5pt]
\textbf{Expert instruction}\\*[2pt]
Stage 1 --- choosing the problem. Topic: \textbf{[Research topic]}\\[4pt]
From where you stand, what is the most important open problem in [Research topic] right now --- the one a good paper would move? State it as a research question and say why it matters and why it is still open. Problem only; no solutions yet.\\[4pt]
Keep it under 900 words.\\[4pt]
\addlinespace[4pt]
\end{longtable}
\endgroup

\begingroup\small
\setlength{\tabcolsep}{0pt}
\renewcommand{\arraystretch}{1.08}
\begin{longtable}{@{}>{\raggedright\arraybackslash}p{\textwidth}@{}}
\toprule
\textbf{Stage 1: anonymous problem shortlist}\\
\midrule\endfirsthead
\toprule
\textbf{Stage 1: anonymous problem shortlist} (continued)\\
\midrule\endhead
\midrule\hfill Continued on next page\\\endfoot
\bottomrule\noalign{\vskip 6pt}
\caption{Stage 1: anonymous problem shortlist}\label{tab:prompt-shortlist}\\
\endlastfoot
\textit{Reading note.} The moderator reads the three problem statements and produces a shortlist. The implementation shuffles the anonymous candidates before voting.\\[5pt]
\textbf{Moderator instruction}\\*[2pt]
From the three statements above, write a shortlist of exactly 3 candidate problems for the discussion to work on. Stay faithful to what was raised --- you may sharpen wording, and combine two problems only where a researcher explicitly connected them. For each, say why it matters: what an answer would change, and who would use it.\\[4pt]
The shortlist is voted on blind, so write each candidate without any reference to who raised it --- no "Expert A", no "one researcher's view". Papers and methods may be named; participants may not.\\[4pt]
Return only this JSON:\\[4pt]
\{\newline \hspace*{0.8em}"candidates": [\newline \hspace*{1.6em}\{"problem": "...", "why\_\allowbreak{}significant": "..."\},\newline \hspace*{1.6em}\{"problem": "...", "why\_\allowbreak{}significant": "..."\},\newline \hspace*{1.6em}\{"problem": "...", "why\_\allowbreak{}significant": "..."\}\newline \hspace*{0.8em}]\newline \}\\[4pt]
\addlinespace[4pt]
\end{longtable}
\endgroup

\begingroup\small
\setlength{\tabcolsep}{0pt}
\renewcommand{\arraystretch}{1.08}
\begin{longtable}{@{}>{\raggedright\arraybackslash}p{\textwidth}@{}}
\toprule
\textbf{Stage 1: blind problem ranking}\\
\midrule\endfirsthead
\toprule
\textbf{Stage 1: blind problem ranking} (continued)\\
\midrule\endhead
\midrule\hfill Continued on next page\\\endfoot
\bottomrule\noalign{\vskip 6pt}
\caption{Stage 1: blind problem ranking}\label{tab:prompt-vote}\\
\endlastfoot
\textit{Reading note.} Turn 2 of problem selection. Each voter receives only the neutral system message, topic and anonymous shortlist, without its expertise prompt, discussion history or other votes.\\[5pt]
\textbf{System message}\\*[2pt]
You are ranking candidate research problems on a topic. Judge them on their own terms; you are not told, and should not guess, who raised which.\\[4pt]
\addlinespace[4pt]
\textbf{User message}\\*[2pt]
Stage 1, turn 2 --- vote. Topic: \textbf{[Research topic]}\\[4pt]
Three candidate problems, shown anonymously in arbitrary order:\\[4pt]
[Candidate problems]\\[4pt]
Rank all three by which one this group should spend the discussion on. Weigh how much the field would gain from an answer, and whether a real paper could plausibly be written on it --- a problem that is easy to measure but changes nothing should not win on tractability alone. One or two sentences of reason per candidate.\\[4pt]
Submit by calling submit\_\allowbreak{}endorsement exactly once.\\[4pt]
\addlinespace[4pt]
\textbf{Tool response (field summary)}\\*[2pt]
Call submit\_\allowbreak{}endorsement once with a ranking array of exactly 3 objects. Each object contains:\newline candidate: integer identifying a candidate (1--3).\newline rank: integer preference rank (1--3), where 1 is most worth pursuing.\newline reason: a string of one or two sentences.\newline Each candidate and rank appears exactly once. All three fields are required; extra fields are rejected.\\[4pt]
\addlinespace[4pt]
\end{longtable}
\endgroup

\begingroup\small
\setlength{\tabcolsep}{0pt}
\renewcommand{\arraystretch}{1.08}
\begin{longtable}{@{}>{\raggedright\arraybackslash}p{\textwidth}@{}}
\toprule
\textbf{Stage 1: final problem selection}\\
\midrule\endfirsthead
\toprule
\textbf{Stage 1: final problem selection} (continued)\\
\midrule\endhead
\midrule\hfill Continued on next page\\\endfoot
\bottomrule\noalign{\vskip 6pt}
\caption{Stage 1: final problem selection}\label{tab:prompt-selection}\\
\endlastfoot
\textit{Reading note.} One of these instructions is sent to the moderator. Vote totals use Borda scores of 3, 2 and 1; only tied top candidates are passed to the tie branch.\\[5pt]
\textbf{Moderator instruction: unique winner}\\*[2pt]
The vote is in. Winning candidate:\newline [Winning problem]\\[4pt]
Tally:\newline [Vote totals]\\[4pt]
Write the problem statement the discussion will work on: 3-5 sentences that state the question, why it matters, and what would count as answering it. Stay with what was raised --- sharpen, don't broaden or merge. Output the problem statement only.\\[4pt]
\addlinespace[4pt]
\textbf{Moderator instruction: tie}\\*[2pt]
The vote is in and tied at the top between:\newline [Tied problems]\\[4pt]
Tally:\newline [Vote totals]\\[4pt]
Choose one of the tied candidates --- only those --- using the reasons the researchers gave, and say in one sentence why.\\[4pt]
Write the problem statement the discussion will work on: 3-5 sentences that state the question, why it matters, and what would count as answering it. Stay with what was raised --- sharpen, don't broaden or merge. Output the problem statement only.\\[4pt]
\addlinespace[4pt]
\end{longtable}
\endgroup

%% file: tables/prompts_outputs.tex
\begingroup\small
\setlength{\tabcolsep}{0pt}
\renewcommand{\arraystretch}{1.08}
\begin{longtable}{@{}>{\raggedright\arraybackslash}p{\textwidth}@{}}
\toprule
\textbf{Shared proposal output format}\\
\midrule\endfirsthead
\toprule
\textbf{Shared proposal output format} (continued)\\
\midrule\endhead
\midrule\hfill Continued on next page\\\endfoot
\bottomrule\noalign{\vskip 6pt}
\caption{Shared proposal output format}\label{tab:prompts-outputs}\\
\endlastfoot
\textit{Reading note.} This exact JSON format is inserted into first-draft, update and final-synthesis instructions. Those tables link here to avoid repeating it. For final synthesis it is wrapped in a one-element array. The expert-attribution field is excluded from pairwise evaluation.\\[5pt]
\textbf{JSON format}\\*[2pt]
\{\newline \hspace*{0.8em}"title": "the idea in one line",\newline \hspace*{0.8em}"problem\_\allowbreak{}and\_\allowbreak{}motivation": "the question this answers and why the field needs it answered",\newline \hspace*{0.8em}"method": "the approach, at the level a paper describes its contribution: what is built, studied or measured, and what makes it different from what exists",\newline \hspace*{0.8em}"experiment": "the primary study that decides the claim --- data, models, baselines, and what result confirms or refutes it --- then, briefly, what you would add if it holds. Keep the experiments simple and feasible, and don't rely on human annotation unless the claim can't be tested without it",\newline \hspace*{0.8em}"novelty": "what is new here",\newline \hspace*{0.8em}"builds\_\allowbreak{}on": "the 2-3 specific results the method actually uses --- each named by paper title and year with what it contributes. Only work the discussion drew on; not background reading",\\[4pt]
\hspace*{0.8em}"distinct\_\allowbreak{}from": "the 1-2 closest existing works and the delta: what this proposal does that each of them does not",\newline \hspace*{0.8em}"potential\_\allowbreak{}impact": "what changes for the field if this works, and what is learned if it does not",\newline \hspace*{0.8em}"trajectory\_\allowbreak{}attribution": \{\newline \hspace*{1.6em}"expert\_\allowbreak{}A": "what this researcher's thinking contributed to the final idea, in a sentence",\newline \hspace*{1.6em}"expert\_\allowbreak{}B": "what this researcher's thinking contributed to the final idea, in a sentence",\newline \hspace*{1.6em}"expert\_\allowbreak{}C": "what this researcher's thinking contributed to the final idea, in a sentence"\newline \hspace*{0.8em}\}\\[4pt]
\}\\[4pt]
\addlinespace[4pt]
\end{longtable}
\endgroup

%% file: tables/prompts_debate.tex
\begingroup\small
\setlength{\tabcolsep}{0pt}
\renewcommand{\arraystretch}{1.08}
\begin{longtable}{@{}>{\raggedright\arraybackslash}p{\textwidth}@{}}
\toprule
\textbf{Stage 2: expert proposals}\\
\midrule\endfirsthead
\toprule
\textbf{Stage 2: expert proposals} (continued)\\
\midrule\endhead
\midrule\hfill Continued on next page\\\endfoot
\bottomrule\noalign{\vskip 6pt}
\caption{Stage 2: expert proposals}\label{tab:prompt-proposal}\\
\endlastfoot
\textit{Reading note.} Turn 1. The task, expert-specific perspective and speech-length rule are shown together as assembled for one expert. Insight name identifies that expert's research line.\\[5pt]
\textbf{Expert instruction}\\*[2pt]
Stage 2, turn 1 --- proposals.\\[4pt]
The problem is settled above. Propose the research you would actually do on it, seen from where you stand. One developed proposal beats several sketches: what is the idea, what is new about it, how would you test it, and what would the field learn either way.\\[4pt]
You are the one here who has spent time inside \textbf{[Insight name]}. Say what that line of work sees in this problem that the other two would miss, and name the tension from it that your proposal resolves. Build on the specific findings you hold --- name the results your proposal depends on. Then bring in whatever else you know.\\[4pt]
Keep it under 900 words.\\[4pt]
\addlinespace[4pt]
\end{longtable}
\endgroup

\begingroup\small
\setlength{\tabcolsep}{0pt}
\renewcommand{\arraystretch}{1.08}
\begin{longtable}{@{}>{\raggedright\arraybackslash}p{\textwidth}@{}}
\toprule
\textbf{Stage 2: first proposal draft}\\
\midrule\endfirsthead
\toprule
\textbf{Stage 2: first proposal draft} (continued)\\
\midrule\endhead
\midrule\hfill Continued on next page\\\endfoot
\bottomrule\noalign{\vskip 6pt}
\caption{Stage 2: first proposal draft}\label{tab:prompt-draft}\\
\endlastfoot
\textit{Reading note.} The moderator reads the three proposals and writes one draft. The cited-paper directory maps evidence identifiers to paper titles and years. Its instruction permits naming only papers actually cited in the discussion.\\[5pt]
\textbf{Moderator instruction}\\*[2pt]
Write the first draft of the proposal from the three proposals above.\\[4pt]
There are three sets of ideas on the table. Don't pick one and reduce the others to controls and caveats, and don't staple all three together. Look for the strongest idea the discussion supports --- often that is a combination, where one researcher's framing gives another's method a reason to exist, or two methods turn out to be one idea seen from two sides --- and build one coherent proposal around it. Where a researcher's ideas genuinely don't fit, leave them out. Merge the ideas, not the experiments: the proposal keeps one primary study that decides the claim, and the other researchers' tests become extensions, not requirements.\\[4pt]
An example of the difference. Three researchers on evaluating long-context models: one wants a benchmark of tasks whose answers require combining distant passages; one has studied how attention concentrates on a few positions; one works on synthetic data generation.\newline \hspace*{0.8em}Bolt-on: "A benchmark of distant-passage tasks, with an attention-concentration analysis as a diagnostic and synthetic data as an ablation."\newline \hspace*{0.8em}Combination: "A benchmark whose tasks are generated synthetically so that the positions of the required passages are controlled, letting it measure directly whether failure tracks the attention-concentration pattern --- turning a benchmark into a test of a hypothesis about why models fail."\newline The second is one idea; the first is three.\\[4pt]
Return one JSON object with this shape:\\[4pt]
\textit{[Shared proposal JSON: Table~\ref{tab:prompts-outputs}]}\\[4pt]
Field values are markdown written for a reader who did not see the discussion: define terms, and don't refer to the researchers anywhere except trajectory\_\allowbreak{}attribution. Papers are cited by title and year, never by [P\#] id --- the reader cannot resolve an id, but a named result is something they can check and build on. This is a draft; it will be revised.\\[4pt]
Return only the JSON object.\\[4pt]
PAPER DIRECTORY --- the papers behind the researchers' [P\#] citations. Use it only to name a cited paper by title and year; papers nobody cited don't belong in the write-up:\newline [Cited-paper directory]\\[4pt]
\addlinespace[4pt]
\end{longtable}
\endgroup

\begingroup\small
\setlength{\tabcolsep}{0pt}
\renewcommand{\arraystretch}{1.08}
\begin{longtable}{@{}>{\raggedright\arraybackslash}p{\textwidth}@{}}
\toprule
\textbf{Stage 2: independent draft review}\\
\midrule\endfirsthead
\toprule
\textbf{Stage 2: independent draft review} (continued)\\
\midrule\endhead
\midrule\hfill Continued on next page\\\endfoot
\bottomrule\noalign{\vskip 6pt}
\caption{Stage 2: independent draft review}\label{tab:prompt-critic}\\
\endlastfoot
\textit{Reading note.} The critic reviews the initial and intermediate drafts. It receives only the current proposal and the latest discussion summary when available, not the debate transcript. The same critic prompt is reused after draft updates.\\[5pt]
\textbf{System message}\\*[2pt]
You are reviewing a research proposal draft cold: you have not seen the discussion that produced it, and you judge only the page.\\[4pt]
Read it as a good reviewer would --- someone who wants this paper to exist and is telling the authors what stands between the draft and that paper. Check:\\[4pt]
(a) Contribution --- is this research, or an engineering adjustment dressed as research? Would the field learn something from the result whichever way it comes out?\newline (b) Soundness --- does the experiment actually test the claim? Could the proposed evidence come out the same whether the idea is right or wrong?\newline (c) Consistency --- do the parts fit together? Does every step have the inputs it needs from an earlier step, and do the sections agree with each other?\newline (d) Feasibility --- could a few people do this in a submission cycle with the compute a university lab can get? Does it build new data or run human annotation where existing resources or automatic measurement would test the same claim?\newline (e) Clarity --- can a reader tell what is proposed, what is claimed, and how they would go about doing it?\\[4pt]
Novelty and importance are not yours to judge here; the discussion handles those. Report at most 6 issues, the ones that most threaten the core claim first, each with the passage it concerns and a concrete suggestion. Submit them by calling submit\_\allowbreak{}critic\_\allowbreak{}report exactly once --- with an empty list if the draft holds up.\\[4pt]
\addlinespace[4pt]
\textbf{User message}\\*[2pt]
Proposal draft under review:\\[4pt]
[Current proposal]\newline [Latest discussion summary, if available]\newline Review it and submit your report with the tool.\\[4pt]
\addlinespace[4pt]
\textbf{Tool response (field summary)}\\*[2pt]
Call submit\_\allowbreak{}critic\_\allowbreak{}report once with an issues array of at most 6 objects, or an empty array if no issues remain. Each issue has four required string fields:\newline location: the affected part of the draft.\newline issue\_\allowbreak{}type: a (Contribution), b (Soundness), c (Consistency), d (Feasibility), or e (Clarity).\newline quote: the passage being questioned.\newline suggestion: a concrete proposed correction.\newline Extra fields are rejected.\\[4pt]
\addlinespace[4pt]
\end{longtable}
\endgroup

\begingroup\small
\setlength{\tabcolsep}{0pt}
\renewcommand{\arraystretch}{1.08}
\begin{longtable}{@{}>{\raggedright\arraybackslash}p{\textwidth}@{}}
\toprule
\textbf{Stage 2: revision and final positions}\\
\midrule\endfirsthead
\toprule
\textbf{Stage 2: revision and final positions} (continued)\\
\midrule\endhead
\midrule\hfill Continued on next page\\\endfoot
\bottomrule\noalign{\vskip 6pt}
\caption{Stage 2: revision and final positions}\label{tab:prompts-stages}\\
\endlastfoot
\textit{Reading note.} The configured three-turn discussion uses turn 2 for revision and turn 3 for final positions. Reviewer notes are empty when no issues remain. Unanswered objections addressed to an expert are also appended when present.\\[5pt]
\textbf{Expert instruction: revision}\\*[2pt]
Stage 2, turn 2 --- improving the draft.\\[4pt]
Current draft:\newline [Current proposal]\newline [Reviewer notes, if any]\newline Answer anything aimed at you. Then push the draft toward a better paper: fix a weakness, strengthen the claim or the way it is tested, bring in something the draft is missing, or --- if you think it is right as it stands --- say so and why. Silence reads as agreement, so say what you think. Substantial changes are welcome if they make it better research.\\[4pt]
Keep it under 900 words.\\[4pt]
\addlinespace[4pt]
\textbf{Expert instruction: final positions}\\*[2pt]
Stage 2, turn 3 --- final positions.\\[4pt]
Current draft:\newline [Current proposal]\newline [Reviewer notes, if any]\newline Settle what is still open --- anything aimed at you, and the reviewer's issues that touch your part of the proposal. Then give your final position: what you would still change, or that the draft is ready. No new directions at this point.\\[4pt]
Keep it under 900 words.\\[4pt]
\addlinespace[4pt]
\end{longtable}
\endgroup

\begingroup\small
\setlength{\tabcolsep}{0pt}
\renewcommand{\arraystretch}{1.08}
\begin{longtable}{@{}>{\raggedright\arraybackslash}p{\textwidth}@{}}
\toprule
\textbf{Stage 2: discussion summary and draft update}\\
\midrule\endfirsthead
\toprule
\textbf{Stage 2: discussion summary and draft update} (continued)\\
\midrule\endhead
\midrule\hfill Continued on next page\\\endfoot
\bottomrule\noalign{\vskip 6pt}
\caption{Stage 2: discussion summary and draft update}\label{tab:prompt-update}\\
\endlastfoot
\textit{Reading note.} After expert responses, the moderator records the discussion and revises the draft. The cited-paper directory maps evidence identifiers to paper titles and years. Its instruction permits naming only papers actually cited in the discussion.\\[5pt]
\textbf{Moderator instruction}\\*[2pt]
Summarize this turn and update the draft. Return one JSON object with exactly two top-level keys:\\[4pt]
\{\newline \hspace*{0.8em}"summary": \{\newline \hspace*{1.6em}"turn\_\allowbreak{}recap": "1-2 sentences",\newline \hspace*{1.6em}"draft\_\allowbreak{}updates": [\{"component": "what changed", "change": "how", "source": "who argued for it"\}],\newline \hspace*{1.6em}"settled": ["points everyone now accepts"],\newline \hspace*{1.6em}"open\_\allowbreak{}attacks": [\{"from": "EXPERT\_\allowbreak{}X", "target": "expert or draft component", "claim": "one sentence", "status": "unanswered"\}],\newline \hspace*{1.6em}"dropped": [\{"component": "what was removed", "reason": "why", "source": "who argued for it"\}],\\[4pt]
\hspace*{1.6em}"unresolved": ["what is still contested"],\newline \hspace*{1.6em}"converged": false\newline \hspace*{0.8em}\},\newline \hspace*{0.8em}"updated\_\allowbreak{}draft": \textit{[Shared proposal JSON: Table~\ref{tab:prompts-outputs}]}\newline \}\\[4pt]
Every change to the draft comes from something a researcher said this turn. Carry forward objections nobody answered; drop the ones that were answered. Set "converged" to true only when no researcher asked for further changes and the reviewer's earlier blocking issues are addressed.\\[4pt]
Field values are markdown written for a reader who did not see the discussion: define terms, and don't refer to the researchers anywhere except trajectory\_\allowbreak{}attribution. Papers are cited by title and year, never by [P\#] id --- the reader cannot resolve an id, but a named result is something they can check and build on.\\[4pt]
Return only the JSON object.\\[4pt]
PAPER DIRECTORY --- the papers behind the researchers' [P\#] citations. Use it only to name a cited paper by title and year; papers nobody cited don't belong in the write-up:\newline [Cited-paper directory]\\[4pt]
\addlinespace[4pt]
\end{longtable}
\endgroup

%% file: tables/prompts_synthesis.tex
\begingroup\small
\setlength{\tabcolsep}{0pt}
\renewcommand{\arraystretch}{1.08}
\begin{longtable}{@{}>{\raggedright\arraybackslash}p{\textwidth}@{}}
\toprule
\textbf{Stage 3: final proposal synthesis}\\
\midrule\endfirsthead
\toprule
\textbf{Stage 3: final proposal synthesis} (continued)\\
\midrule\endhead
\midrule\hfill Continued on next page\\\endfoot
\bottomrule\noalign{\vskip 6pt}
\caption{Stage 3: final proposal synthesis}\label{tab:prompt-synthesis}\\
\endlastfoot
\textit{Reading note.} The moderator receives the substantive discussion and latest draft, and returns one final proposal. The cited-paper directory maps evidence identifiers to paper titles and years. Its instruction permits naming only papers actually cited in the discussion.\\[5pt]
\textbf{Moderator instruction}\\*[2pt]
Stage 3 --- final proposal. From the discussion and the latest draft, write the proposal a reader who never saw the debate would receive. Apply the changes the group agreed on. Where a reviewer's issue was never resolved, fix it if the discussion gives you the means, otherwise state it as a limitation. Keep the problem from Stage 1. The method or experiment must use at least one specific recent result the researchers brought --- used, not mentioned: some step of the proposal depends on it being true.\\[4pt]
Return a JSON array holding exactly one proposal object:\\[4pt]
\textit{[Shared proposal JSON: Table~\ref{tab:prompts-outputs}]}\\[4pt]
Field values are markdown written for a reader who did not see the discussion: define terms, and don't refer to the researchers anywhere except trajectory\_\allowbreak{}attribution. Papers are cited by title and year, never by [P\#] id --- the reader cannot resolve an id, but a named result is something they can check and build on.\\[4pt]
Return only the JSON array.\\[4pt]
PAPER DIRECTORY --- the papers behind the researchers' [P\#] citations. Use it only to name a cited paper by title and year; papers nobody cited don't belong in the write-up:\newline [Cited-paper directory]\\[4pt]
\addlinespace[4pt]
\end{longtable}
\endgroup

%% file: tables/prompts_evaluation.tex
\begingroup\small
\setlength{\tabcolsep}{0pt}
\renewcommand{\arraystretch}{1.08}
\begin{longtable}{@{}>{\raggedright\arraybackslash}p{\textwidth}@{}}
\toprule
\textbf{Pairwise research idea evaluation}\\
\midrule\endfirsthead
\toprule
\textbf{Pairwise research idea evaluation} (continued)\\
\midrule\endhead
\midrule\hfill Continued on next page\\\endfoot
\bottomrule\noalign{\vskip 6pt}
\caption{Pairwise research idea evaluation}\label{tab:prompts-evaluation}\\
\endlastfoot
\textit{Reading note.} Proposal 0 and Proposal 1 are the final ideas in the current presentation order; they are exchanged for the reverse-order judgment. The five criteria appear directly in the prompt. Effectiveness is reported as expected effectiveness in the paper. Braced descriptions in the response format specify required outputs, not missing input values.\\[5pt]
\textbf{User message}\\*[2pt]
You are a judge in a competition. You have to decide which idea is better.\\[4pt]
The idea0 is: [Proposal 0]\\[4pt]
The idea1 is: [Proposal 1]\\[4pt]
The topic is: [Research topic]\\[4pt]
Which idea do you think is better? Judge each criterion separately.\\[4pt]
Here are your evaluation criteria:\newline 1. Novelty: Which idea makes the more substantive original intellectual contribution: an insight, hypothesis, problem formulation, method, or explanatory connection? A meaningful combination of familiar techniques can be original when their interaction enables new understanding or capability. Mere assembly, complexity or renaming is not novelty. Do not favor or penalize an idea because it is easier to implement. Support duplication claims with identifiable prior work when known; acknowledge uncertainty rather than inventing references.\newline 2. Significance: Which central contribution, if established, would teach the field more or address a more consequential bottleneck? Consider changes in understanding, useful capabilities and what others could build on, including focused studies and informative negative results. Judge supported potential value independently of execution difficulty. Breadth and ambitious language alone do not establish significance.\newline 3. Feasibility: Which idea has the more credible path for an academic group to implement its core mechanism and test its central claim? Distinguish essential requirements from optional extensions and manageable adjustments from fundamental barriers. Adjustments must preserve the intellectual contribution; do not invent missing solutions or weaken the claim. Lower cost or shorter duration alone should not decide between two credible plans. Compare essential resources, assumptions and unresolved dependencies; novelty and ambition do not excuse an infeasible plan.\newline 4. Clarity: Which idea more clearly connects its central contribution, mechanism, assumptions and proposed test? Require the details needed to assess that connection, not an exhaustive implementation recipe. Do not favor simpler scope or a longer checklist of datasets and metrics over a precisely explained intellectual contribution. Penalize undefined concepts and missing reasoning, not conceptual ambition or routine implementation choices left open.\newline 5. Effectiveness: Which idea's mechanism and evaluation more credibly deliver its stated research objective and produce justified knowledge? A hypothesis test can be effective even when it refutes the hypothesis; a claimed improvement still needs a plausible mechanism for achieving it. Prefer tests that distinguish the claim from credible alternatives and support the conclusions actually sought. Do not substitute ease of execution or guaranteed positive results for evidential strength. Penalize circular evaluation, confounding and unsupported inferential leaps.\\[4pt]
General rules for all criteria (very important):\newline - Judge CONTENT, not presentation. Length, number of sections, bullet density, formatting polish, and confident tone are NOT evidence of quality on any criterion. Prefer the idea that is specific and straight-to-the-point over one that is longer or more elaborately formatted.\newline - Do not reward claims you cannot verify. Citations, numbers, or assertions about prior work that you cannot check count neither for nor against the idea; they never raise novelty or effectiveness on their own.\newline - Keep the criteria orthogonal. Judge each criterion only against its own definition: do not lower feasibility because an idea is unoriginal, and do not raise novelty because an idea is well written.\newline - Every idea has limitations. Only limitations that bear on the criterion you are currently judging should affect that criterion.\newline - State of the art is not required. Impact may come from efficiency, generalizability, interpretability, robustness, or fairness, not only from beating a leaderboard.\\[4pt]
Note:\newline Avoid any position biases and ensure that the order in which the responses were presented does not influence your decision. DO NOT allow the LENGTH of the responses to influence your evaluation, choose the one that is straight-to-the-point instead of unnecessarily verbose. Be as objective as possible. (very important!!!)\\[4pt]
If you think idea0 is better than idea1, you should output 0. If you think idea1 is better than idea0, you should output 1. If you think idea0 and idea1 are equally good, you should output 2. Only output 2 when the two ideas are genuinely equivalent on that criterion - do not use it to avoid making a decision.\\[4pt]
Your output should be strictly in following format:\newline Your thinking process:\newline ...\\[4pt]
Your choice:\newline \textless{}novelty\_\allowbreak{}reason\textgreater{}\{ 1-2 sentences citing specific content of the idea \}\textless{}/novelty\_\allowbreak{}reason\textgreater{}\newline \textless{}novelty\textgreater{}\{ Your choice (0, 1, or 2) for novelty \}\textless{}/novelty\textgreater{}\newline \textless{}significance\_\allowbreak{}reason\textgreater{}\{ 1-2 sentences citing specific content of the idea \}\textless{}/significance\_\allowbreak{}reason\textgreater{}\newline \textless{}significance\textgreater{}\{ Your choice (0, 1, or 2) for significance \}\textless{}/significance\textgreater{}\newline \textless{}feasibility\_\allowbreak{}reason\textgreater{}\{ 1-2 sentences citing specific content of the idea \}\textless{}/feasibility\_\allowbreak{}reason\textgreater{}\newline \textless{}feasibility\textgreater{}\{ Your choice (0, 1, or 2) for feasibility \}\textless{}/feasibility\textgreater{}\\[4pt]
\textless{}clarity\_\allowbreak{}reason\textgreater{}\{ 1-2 sentences citing specific content of the idea \}\textless{}/clarity\_\allowbreak{}reason\textgreater{}\newline \textless{}clarity\textgreater{}\{ Your choice (0, 1, or 2) for clarity \}\textless{}/clarity\textgreater{}\newline \textless{}effectiveness\_\allowbreak{}reason\textgreater{}\{ 1-2 sentences citing specific content of the idea \}\textless{}/effectiveness\_\allowbreak{}reason\textgreater{}\newline \textless{}effectiveness\textgreater{}\{ Your choice (0, 1, or 2) for effectiveness \}\textless{}/effectiveness\textgreater{}\\[4pt]
\addlinespace[4pt]
\end{longtable}
\endgroup